\documentclass{article} % For LaTeX2e
\usepackage{collas2022_conference,times}

\usepackage[utf8]{inputenc} % allow utf-8 input
\usepackage[T1]{fontenc}    % use 8-bit T1 fonts
\usepackage{url}            % simple URL typesetting
\usepackage{booktabs}       % professional-quality tables
\usepackage{amsfonts}       % blackboard math symbols
\usepackage{nicefrac}       % compact symbols for 1/2, etc.
\usepackage{microtype}      % microtypography
\usepackage{amsmath}
\usepackage{amsthm}
\usepackage{graphicx}

\usepackage{xcolor}
\definecolor{dkgreen}{rgb}{0,0.6,0}

\usepackage{algorithm}
\usepackage{forloop}
\usepackage{algpseudocode}

\usepackage{xspace}
\usepackage{overpic}

\usepackage{amssymb}
\usepackage{pifont}

\usepackage{caption}
\usepackage{subcaption}
\usepackage{wrapfig} % figure next to text
\usepackage{enumitem} % no indent for list
\usepackage{enumerate} %

\newtheorem{proposition}{Proposition}[section]

\newcommand{\boldy}{\mathbf{y}}
\newcommand{\boldR}{\mathbf{R}}

\newcommand{\boldtheta}{\boldsymbol{\theta}}
\newcommand{\boldomega}{\boldsymbol{\omega}}
\newcommand{\boldphi}{\boldsymbol{\phi}}
\newcommand{\boldx}{\mathbf{x}}
\newcommand{\boldA}{\mathbf{A}}

\newcommand{\boldC}{\mathbf{C}}

\newcommand{\boldI}{\mathbf{I}}
\newcommand{\boldW}{\mathbf{W}}

\newcommand{\bolds}{\mathbf{s}}
\newcommand{\boldn}{\mathbf{n}}
\newcommand{\boldz}{\mathbf{z}}

\newcommand{\oBase}{\textit{Base}\xspace}
\newcommand{\oMI}{\textit{Base+MI}\xspace}
\newcommand{\oCMI}{\textit{Base+CMI}\xspace}

\newcommand{\wBase}{\textit{``Base''}\xspace}
\newcommand{\wMI}{\textit{``Base+MI''}\xspace}
\newcommand{\wCMI}{\textit{``Base+CMI''}\xspace}

\usepackage{amsmath,amsfonts,bm}

\def\eqref#1{equation~\ref{#1}}
\def\1{\bm{1}}

\DeclareMathAlphabet{\mathsfit}{\encodingdefault}{\sfdefault}{m}{sl}
\SetMathAlphabet{\mathsfit}{bold}{\encodingdefault}{\sfdefault}{bx}{n}

\newcommand{\Var}{\mathrm{Var}}

\DeclareMathOperator{\Tr}{Tr}

\usepackage{hyperref}
\hypersetup{
    colorlinks=true,
    linkcolor=red,
    filecolor=magenta,      
    urlcolor=blue,
    citecolor=purple,
    pdftitle={Overleaf Example},
    pdfpagemode=FullScreen,
}

\title{Disentanglement and Generalization Under Correlation Shifts}

\author{Christina M. Funke\thanks{Equal contribution. \textsuperscript{$\dagger$}Shared senior authors.}* \\
University of T\"ubingen \\
\And Paul Vicol* \\University of Toronto\\ Vector Institute
\And Kuan-Chieh Wang \\University of Toronto\\ Vector Institute
\AND Matthias Kümmerer\textsuperscript{$\dagger$} \\University of T\"ubingen
\And Richard Zemel\textsuperscript{$\dagger$} \\University of Toronto\\ Vector Institute
\And Matthias Bethge\textsuperscript{$\dagger$} \\University of T\"ubingen
}

\collasfinalcopy

\begin{document}

\maketitle

\begin{abstract}
\vspace{-0.4cm}
Correlations between factors of variation are prevalent in real-world data. Exploiting such correlations may increase predictive performance on noisy data; however, often correlations are not robust (e.g., they may change between domains, datasets, or applications) and models that exploit them do not generalize when correlations shift. Disentanglement methods aim to learn representations which capture different factors of variation in latent subspaces. A common approach involves minimizing the mutual information between latent subspaces, such that each encodes a single underlying attribute. However, this fails when attributes are correlated. We solve this problem by enforcing independence between subspaces conditioned on the available attributes, which allows us to remove only dependencies that are not due to the correlation structure present in the training data. We achieve this via an adversarial approach to minimize the conditional mutual information (CMI) between subspaces with respect to categorical variables. We first show theoretically that CMI minimization is a good objective for robust disentanglement on linear problems. We then apply our method on real-world datasets based on MNIST and CelebA, and show that it yields models that are disentangled and robust under correlation shift, including in weakly supervised settings.
\end{abstract}

\section{Introduction}
\label{sec:introduction}
\vspace{-0.2cm}
Disentangled representations can be useful for improving fairness~\citep{locatello2019fairness}, interpretability~\citep{adel2018discovering}, controllable generative modeling~\citep{he2019attgan}, and transfer to downstream tasks~\citep{van2019disentangled}.
In addition, they can improve robustness on out-of-distribution data~\citep{higgins2017darla} (e.g., for domain adaptation~\citep{ilse2020diva} and domain generalization~\citep{ben2009robust}).
Most research on disentanglement has assumed that the underlying factors of variation in the data are \textit{independent} (e.g., that factors are not correlated).
However, this assumption is often violated in real-world settings: for example, in domain adaptation, the class distribution often shifts between domains (yielding a correlation between the class and domain); 
in natural images, there is often a strong correlation between the foreground and background~\citep{beery2018recognition}, or between multiple foreground objects that tend to co-occur (e.g., a keyboard and monitor)~\citep{tsipras2020imagenet, beyer2020we}.
Importantly, correlated data occur in areas that affect people's lives, including in healthcare~\citep{chartsias2018factorised} and fairness applications~\citep{madras2018learning,creager2019flexibly,locatello2019fairness}, and correlation shifts in these applications are common (e.g., demographics are likely to differ from one hospital to another).

The goal of disentanglement is to encode data into independent subspaces that preferably match the ground truth generative factors.
A common approach to achieve this (used in ICA, PCA, and VAEs) is to ensure that the latent subspaces share as little information as possible, by minimizing the mutual information (MI) between subspaces.
However, recently it has been shown that this fails to disentangle correlated factors~\citep{trauble2020independence}.
Several works have sought to address this by introducing partial supervision~\citep{trauble2020independence,shu2019weakly,locatello2020weakly}.
Here, we show that even with \textit{full} supervision, minimizing the MI can fail: it is impossible to encode generative factors into independent subspaces if they are correlated in the training data.
To address this, we propose minimizing the MI between subspaces \textit{conditioned} on the correlated attributes.

We compare three objective functions for learning disentangled representations: 1) standard supervised losses (such as mean-squared error or cross-entropy) that encourage each subspace to encode a specific attribute; 2) a supervised loss plus \textit{unconditional} MI minimization; and 3) a supervised loss plus \textit{conditional} MI (CMI) minimization.
We first show that approaches (1) and (2) fail on correlated and noisy data: minimizing a supervised loss cannot enforce that there is little information shared between subspaces; MI minimization is too strong a constraint to satisfy when the underlying factors of variation are correlated, and thus minimizing MI leads to decreased performance.
We then show that minimizing CMI yields disentangled representations that are robust to correlation shifts.

Overall, we aim to establish conditional independence as the correct notion of independence between latent subspaces when disentangling data with correlated factors of variation.

\paragraph{Contributions.}
\begin{itemize}
\vspace{-0.1cm}
    \setlength\itemsep{-0.1em}
    \item Most disentanglement metrics used in the literature assume that the attributes are uncorrelated, and thus are not directly applicable to correlated data.
    We propose to use the \textit{predictive performance under correlation shift} as a \textit{measure of disentanglement} applicable to settings with correlated factors of variation.
    \item We analyze the behavior of each objective function on a linear regression problem where all quantities of interest can be computed analytically (Section~\ref{sec:linear-reg}). We show that minimizing the CMI between latent subspaces yields a solution robust to test-time correlation shifts, while minimizing the unconditional MI (or only a supervised loss) does not.
    \item We describe an adversarial approach for learning conditionally disentangled representations (Section~\ref{sec:method}).
    \item Then, we apply our approach to CMI minimization to two tasks based on real-world datasets---a multi-digit occluded MNIST task and correlated CelebA---and demonstrate improved performance under correlation shift relative to baselines (Section~\ref{sec:experiments}).
    \item We investigate the interplay between correlation strength and noise level in the training data. When data are noisy and have strong correlations, the noise forces the model to rely on correlations when making a prediction; this leads to failures of the baseline approaches when correlations shift at test-time, and demonstrates the benefits of CMI minimization, which performs well across correlation strengths and noise levels.
    \item We show that CMI minimization can be applied in the weakly supervised setting, and show significant gains compared to baselines.
\end{itemize}
\vspace{-0.2cm}

Our code is available \href{https://github.com/asteroidhouse/conditional-disentanglement}{on Github}.

\section{Background \& Related Work}
\label{sec:related-work}
\vspace{-0.2cm}

\paragraph{ICA/ISA.}
Disentanglement is related to blind source separation (BSS), as both problems revolve around the question of identifiability.
A classic approach to BSS is Independent Component Analysis (ICA)~\citep{comon1994independent,jutten1991blind,bell1997independent,olshausen1996emergence}, which assumes statistical independence between the source variables~\citep{jutten1991blind,jutten2003advances}.
Independent Subspace Analysis (ISA)~\citep{hyvarinen2000emergence}, or multidimensional ICA~\citep{cardoso1998multidimensional}, is a generalization of ICA where each component is a $k$-dimensional subspace; dimensions within a subspace may have dependencies, while dimensions from different subspaces must be independent.
Our work can be seen as a form of nonlinear ISA that enforces conditional independence between subspaces.

\vspace{-0.1cm}
\paragraph{Correlations Between Features.}
With roots in ICA, most research on disentanglement focuses on data that was generated by independent factors, including synthetic benchmarks such as dSprites~\citep{dsprites17}, Shapes3D~\citep{3dshapes18}, Cars3D~\citep{reed2015deep}, SmallNORB~\citep{lecun2004learning}, or MPI3D~\citep{gondal2019transfer}.
In real-world datasets on the other hand, factors are often correlated~\citep{WelinderEtal2010,lin2014microsoft}. 
~\cite{trauble2020independence} pointed out the challenges that arise when attempting to learn disentangled representations on correlated data, and performed a large-scale empirical evaluation of the effect of correlations on widely-used VAE-based disentanglement models.
They proposed two approaches to ameliorate the harmful effects of correlations: 1) introducing weak supervision during training, and 2) labeling data post-hoc to ``correct'' a pre-trained encoder. 
We show that even with full supervision, correlations are problematic when enforcing independence between latent subspaces.
Causally-informed modeling~\citep{zhang2020causal} is another approach to learning disentangled representations and extracting invariant features.
To investigate the effect of correlations systematically, it is common to modify existing datasets to induce correlations, for example by subsampling the data, or generating synthetic datasets with the desired properties~\citep{dittadi2020transfer, cimpoi2014describing, jacobsen2018excessive, locatello2019challenging}.
We follow this approach in our experiments.

\vspace{-0.1cm}
\paragraph{Unsupervised and Weakly-Supervised Disentanglement.}
Disentangled representation learning is often studied in the unsupervised setting, where the ground-truth factors of variation are unknown.
Widely-used approaches for this include variational autoencoders (VAEs)~\citep{kingma2013auto} and their variants (beta-VAE~\citep{higgins2016beta}, TC-beta-VAE~\citep{chen2018isolating}, FactorVAE~\citep{kim2018disentangling}, etc.).
However, it was shown by \cite{locatello2019challenging} that the assumption of independent source variables (e.g., attributes) is questionable, and that \textit{purely unsupervised} disentanglement may not be possible.
This spurred interest in \textit{weakly-supervised} methods~\citep{shu2019weakly,locatello2020weakly}, where weak supervision is provided in the form of partial labels or grouping information~\citep{bouchacourt2018multi,nemeth2020adversarial, klindt2020towards}.
In this paper, we focus on comparing MI and CMI minimization in the fully-supervised setting, as this is already challenging and provides useful insights.

\vspace{-0.1cm}
\paragraph{Domain Adaptation/Generalization.}
We use predictive performance under correlation shift as a measure for the quality of disentanglement. This is closely related to the fields of domain adaptation and generalization, with the difference that we assume access to one source domain only. The goal of most related work in this field is to learn representations from multiple source domains that transfer to known (e.g., adaptation) or previously unseen (e.g., generalization) target domains.
This is done by either learning domain-invariant representations which discard domain information~\citep{tzeng2017adversarial} or by learning disentangled representations, with latent subspaces that correspond to the domain and the class, respectively~\citep{peng2019domain,ilse2020diva,liu2018unified}.
For the latter approach, disentanglement is achieved by minimizing the mutual information between latent subspaces~\citep{cheng2020club, gholami2020unsupervised,nemeth2020adversarial}.
~\cite{zhao2019learning} discuss fundamental problems inherent in learning domain-invariant representations when there are correlations between classes and domains (e.g., when the class distribution shifts in the target domain).
The goal of Invariant Risk Minimization \citep{arjovsky2019invariant} is to find correlations that are invariant over multiple training domains in order to improve generalization to out-of-distribution data.

\vspace{-0.1cm}
\paragraph{Fairness.}
An important application of disentanglement is fairness.
As machine learning systems are typically trained on historical data, they often inherit past biases (e.g., from human decision-makers).
This may result in unfair treatment on the basis of sensitive properties such as ethnicity, gender, or disability.
Typically, this can be addressed by modifying the training data to be unbiased or by adding a regularizer (e.g. based on mutual information) that quantifies and minimizes the degree of bias~\citep{kamiran2009classifying, kamishima2011fairness, zemel2013learning, hardt2016equality, cho2020fair}.

\vspace{-0.1cm}
\paragraph{Mutual Information.}
The mutual information (MI) between two random variables $\boldx$ and $\boldy$, denoted $I(\boldx ; \boldy)$, is the KL divergence between the joint distribution $p(\boldx,\boldy)$ and the product of the marginal distributions $p(\boldx) p(\boldy)$: $I(\boldx ; \boldy) = D_\text{KL}[p(\boldx,\boldy) || p(\boldx) p(\boldy)]$.
Minimization of MI has been used to implement an information bottleneck~\citep{alemi2016deep} and to factorize representations~\citep{jacobsen2018excessive}.
MI minimization is at the heart of many approaches to disentanglement.
The \textit{conditional mutual information} (CMI) is defined as:
$I(\boldx; \boldy \mid \boldz) = \mathbb{E}_{\boldz} \left[ D_\text{KL}[p(\boldx, \boldy \mid \boldz)\ ||\ p(\boldx \mid \boldz) p(\boldy \mid \boldz) \right]$.
CMI measures the dependency between two variables given that we know the value of a third variable.
For example, there is a dependency between a country's number of Nobel laureates per capita and chocolate consumption per capita~\citep{prinzChocolateConsumptionNoble2020}.
However, this dependency is largely explained by the wealth of a country, thus $I(nobel; chocolate \mid wealth) < I(nobel; chocolate)$.
In general, the CMI can be smaller or larger than the unconditional MI.

\vspace{-0.1cm}
\paragraph{Estimating \& Optimizing Mutual Information.}
Many approaches have been proposed for MI and CMI estimation and optimization.
The Mutual Information Neural Estimator (MINE)~\citep{belghazi2018mine} uses a lower-bound of the MI based on the Donsker-Varadhan dual representation of the KL divergence~\citep{donsker1983asymptotic}.
~\cite{poole2019variational} provide an overview of variational bounds that can be used to estimate MI; most are \textit{lower bounds}, which are useful in principle for \textit{maximizing} MI, but which have also been used to minimize MI (even though minimizing a lower bound is not guaranteed to decrease MI).
CLUB~\citep{cheng2020club} introduced a variational upper bound of MI, providing a more principled objective for minimizing MI.
Several CMI estimators have been proposed, including conditional-MINE~\citep{molavipour2020conditional}, C-MI-GAN~\citep{mondal2020c}, CCMI~\citep{mukherjee2020ccmi}, and an approach based on nearest neighbors~\citep{molavipour2020neural}.
Many approaches to MI minimization are based on batchwise shuffling of latent subspaces, sometimes referred to as metameric sampling~\citep{belghazi2018mine,nemeth2020adversarial,feng2018dual,park2020swapping,peng2019domain}.
The approach we use in Section~\ref{sec:method} follows this paradigm of latent-space shuffling.

\section{Disentanglement with Correlated Variables: Motivating CMI}
\label{sec:linear-reg}

A summary of notation is provided in Appendix~\ref{app:notation}.

\paragraph{Problem Statement.}
Suppose we observe noisy data $\boldx \in \mathbb{R}^m$ obtained from an (unknown) generative process $\boldx=g(\bolds)$ where $\bolds = (s_1, s_2, \dots, s_K)$ are the \textit{underlying factors of variation}, also called source variables or attributes, which may be correlated with each other.
We wish to find a mapping $f : \mathbb{R}^m \to \mathbb{R}^n$ to a latent space $f(\boldx) = \boldz = (\boldz_1, \boldz_2, \dots, \boldz_K)$ such that each attribute $s_k$ can be recovered from the corresponding latent subspace $\boldz_k$ by a linear mapping $\boldR_k$, e.g., $\hat{s}_k = \boldR_k \boldz_k$ such that $\hat{s}_k \approx s_k$.
We denote by $\boldz_{-i}$ the set of subspaces $\{ \boldz_1, \dots, \boldz_{i-1}, \boldz_{i+1}, \dots, \boldz_K \}$.
We consider three different objectives for learning the latent subspaces: 1) minimizing a supervised loss $L$ (e.g., mean squared error or cross-entropy), $\sum_{i=1}^K L(\hat{s}_i, s_i)$, denoted \wBase; 2) minimizing the \textit{unconditional mutual information between subspaces} in addition to the supervised loss, $\sum_i L(\hat{s}_i, s_i) + I(\boldz_1, \dots, \boldz_K)$, denoted \wMI; and 3) minimizing the \textit{conditional mutual information between subspaces conditioned on observed attributes}, in addition to the supervised loss, 
$\sum_i L(\hat{s}_i, s_i) + I(\boldz_i ; \boldz_{-i} \mid s_i)$
denoted \wCMI.
We wish to learn a model that is robust to correlation shifts, e.g., if we train on data where $\text{corr}(s_i, s_j) > 0$, then we desire that the resulting model will perform similarly on uncorrelated data, $\text{corr}(s_i, s_j) = 0$, or anticorrelated data, $\text{corr}(s_i, s_j) < 0$.

In this section, we motivate the use of CMI minimization for learning robust disentangled representations.
We use a linear regression task that can be solved analytically, and for which all quantities of interest, including MI and CMI, can be computed in closed form.
This allows us to compare the solutions obtained via the vanilla mean-squared error objective (\oBase) to the solutions obtained by minimizing the MSE \textit{under the constraint} that the MI or CMI between latent subspaces is minimized.
This yields insight into the behavior of the objectives in the idealized case where the constraints they prescribe ($I(z_1 ; z_2) = 0$ for MI or $I(z_1 ; z_2 \mid s_1) = I(z_1 ; z_2 \mid s_2) = 0$ for CMI) are exactly satisfied.

First, we show that the supervised loss alone does not yield robust disentangled representations.
Then, we show that additionally minimizing the unconditional MI forces the model to learn an \textit{even worse solution}.
Finally, we show that minimizing the conditional MI yields appropriately disentangled representations that are robust to correlation shift.

\begin{table*}
\centering
\begin{tabular}{l   c  c  c}
\toprule
     & \textbf{Base} & \textbf{Base \texttt{+} MI} & \textbf{Base \texttt{+} CMI}\\
\midrule
     \textbf{Variance Explained, Training (Corr = 0.8)} & 91.9\% & 69.8\% & 90.9\%\\ 
     \textbf{Variance Explained, Test (Corr = 0)} & 87.6\% & 65.0\% & 90.9\%\\
    \textbf{Regression Matrix $M$} (where $\hat{\bolds} = M \boldx$)
      &$\begin{pmatrix}0.81 & 0.14 \\ 0.14 & 0.81\end{pmatrix}$
      &$\begin{pmatrix}1.07 & -0.46 \\ -0.46 & 1.07\end{pmatrix}$
      &$\begin{pmatrix}1 & 0 \\ 0 & 1\end{pmatrix}$\\
\bottomrule
\end{tabular}
\vspace{-0.2cm}
\caption{\small Robustness of linear regression under correlation shift for each of the objectives \oBase, \oMI, and \oCMI.
Here, the observations and predictions are in $\mathbb{R}^2$.
The performance of the \oBase model drops under correlation shift.
The optimal solution under the constraint of minimal MI, $I(z_1 ; z_2) = 0$, fails to model the in-distribution correlated training data.
The solution with minimal \textit{conditional} MI, $I(z_1 ; z_2 \mid s_1) = I(z_1 ; z_2 \mid s_2) = 0$, maintains consistent performance under correlation shift.
Note that because the generative process is given by $g(\bolds) = \boldA \bolds = \boldI \bolds$, the inverse is $\boldA^{-1} = \boldI$.
In the last row, we see that only Base + CMI recovers this true inverse.
}
\vspace{-0.4cm}
\label{table:regression}
\end{table*}

\subsection{Full Supervision Does Not Yield Disentanglement}
\label{sec:supervised-disentanglement}

\begin{figure}
\begin{subfigure}{\textwidth}
    \centering
    \includegraphics[width=1\linewidth]{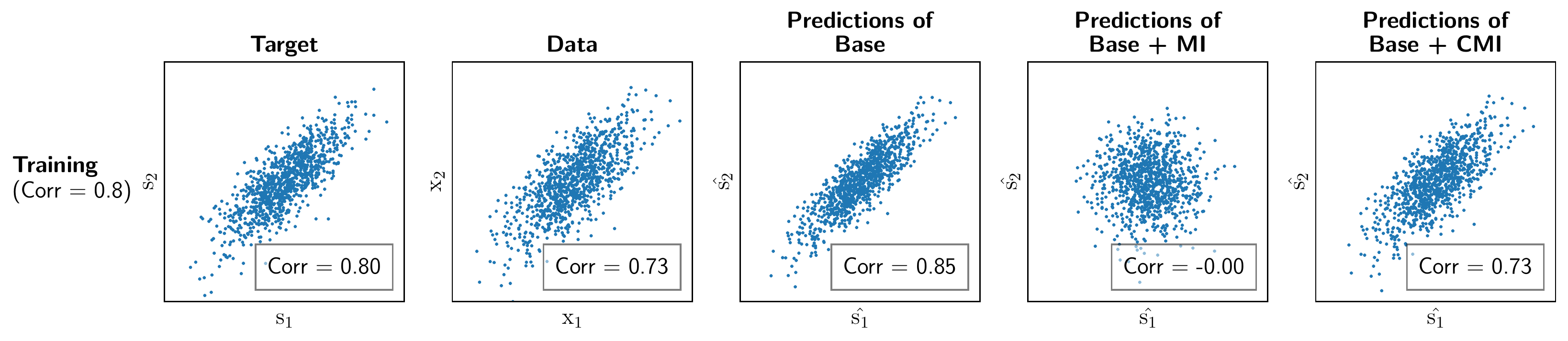}
\end{subfigure}%
\hfill
\begin{subfigure}{\textwidth}
    \centering
    \includegraphics[width=1\linewidth]{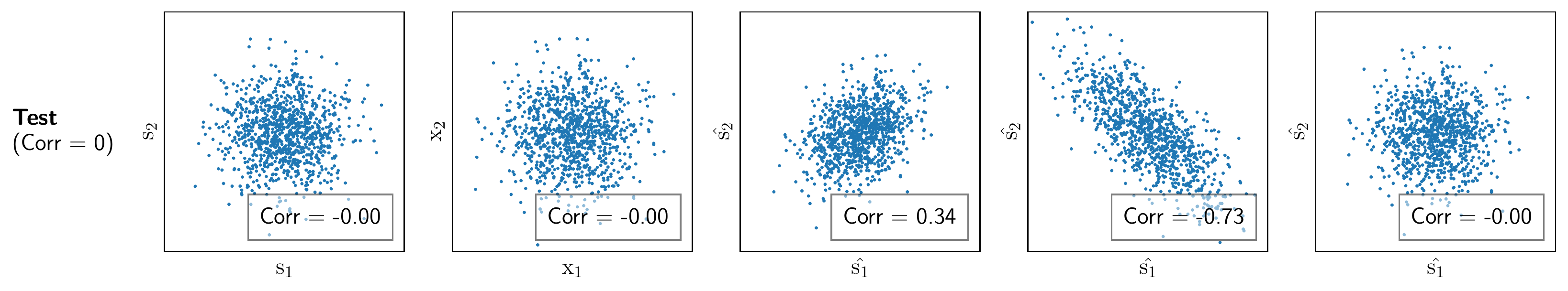}
\end{subfigure}
\vspace{-0.4cm}
\caption{\small \textbf{Visualisation of targets $\mathbf{s}$, input data $\mathbf{x}$ and the predictions $\hat{\mathbf{s}}$ made by models using each of the different objectives \{\oBase, \oMI, \oCMI\}.} For \oBase, the predictions are more correlated than the data, revealing that the correlation in the training data is used to compensate for the noise. \oMI leads to uncorrelated predictions. This cannot be the correct solution, as the targets are correlated. Only for \oCMI does the correlation between the predictions and data match for both training and test data.}
\vspace{-0.3cm}
\label{fig:linear_correlation_of_predictions}
\end{figure}

Here, we introduce a linear regression problem with correlated attributes.
First, we analyze the solution obtained by optimizing only the \oBase objective, which in this case is the mean squared error.
Consider a linear generative model with correlated Gaussian source variables $\bolds$, given by:
\begin{align*} 
\mathbf{x} &= \mathbf{A} \mathbf{s} + \mathbf{n} \quad,\quad
\mathbf{s} \sim \mathcal{N}(\mathbf{0}, \mathbf{C_s}) \quad,\quad  \mathbf{n} \sim \mathcal{N}(\mathbf{0}, \mathbf{C_n})
\end{align*}
where $\mathbf{A}$ is the ground-truth mixing matrix and $\mathbf{C_s}$ and $\mathbf{C_n}$ are the covariance matrices for the source and noise variables, respectively.
We assume that $\mathbf{x}$ is observed and wish to disentangle the underlying source variables $\mathbf{s}$; this corresponds to finding the mapping $\mathbf{A}^{-1}$ that inverts the data generating process.
When we have access to the source variables, a natural approach is to minimize a supervised loss to ensure that each subspace contains information about its attribute.
The optimal linear regression solution, both in the least squares sense and with respect to maximum likelihood, is given by the posterior mean:
\begin{align} 
\hat{\mathbf{s}}(\mathbf{x}) &= \mathbb{E}\left[\mathbf{s}\mid\mathbf{x}\right] = \mathbf{C}_{\mathbf{sx}}\mathbf{C}_{\mathbf{x}}^{-1}\mathbf{x} \label{eq:full-sup}
\end{align}
where $\mathbf{C_{sx}}$ and $\mathbf{C_x}$ are the following covariance matrices:
\begin{align}
\boldC_{\bolds \boldx} &= \mathbb{E} \left[\bolds( \boldA \bolds + \boldn)^\top \right] = \boldC_{\bolds} \boldA^\top \\
\boldC_{\boldx} &= \boldA \boldC_{\bolds} \boldA^\top + \boldC_{\boldn} 
\end{align}
The least-squares optimal mapping $\boldC_{\bolds \boldx} \boldC_{\boldx}^{-1}$ in Eq.~\ref{eq:full-sup} is not equal to the inverse $\boldA^{-1}$ of the generative model, as it is biased by the correlation structure $\mathbf{C_s}$ and $\mathbf{C_n}$ towards directions of maximal signal-to-noise ratio.
Thus, regression is sensitive to noise, and this can lead to failures when evaluating the model on correlation-shifted data.
For this Gaussian problem, we can compute the expected mean squared error (and therefore the expected variance explained) analytically:
\begin{equation}
\mathbb{E}\left[ (\mathbf{s}-\hat{\mathbf{s}}(\boldx))^2 \right] = \Var \left( \mathbf{s} \right) = \Tr( \boldC_{\mathbf{s}})
\end{equation}
In Table~\ref{table:regression}, we see that in the two-dimensional case where $\bolds=(s_1, s_2)$ for $\boldA=\boldI$, $\mathbf{C_n}=0.01
\cdot\boldI$ and the train-time correlation is $\text{corr}(s_1, s_2)=0.8$, $\mathbf{\hat s}$ explains $91.9\%$ of the variance in $\bolds$ (column \wBase).
However, when the correlation between $s_1$ and $s_2$ shifts at test time, such that $\text{corr}(s_1, s_2) = 0$, then performance drops to $87.6\%$.
This drop occurs because the estimator $\mathbf{\hat s}$ tries to make use of the assumed correlation between $s_1$ and $s_2$ to counteract the information lost due to noise, but this correlation is no longer present in the test data (see also Figure~\ref{fig:linear_correlation_of_predictions}).
The gap in performance between correlated and uncorrelated data indicates that $s_1$ and $s_2$ have not been correctly disentangled.

\begin{figure}
    \begin{subfigure}{0.7\textwidth}
    \includegraphics[width=\linewidth]{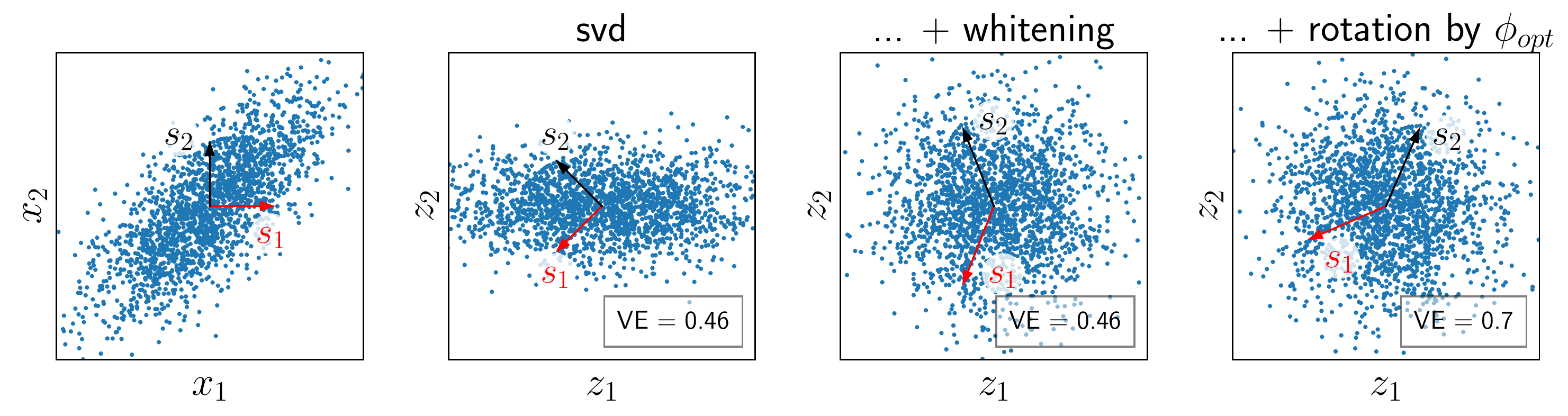}
    \end{subfigure}%
    \begin{subfigure}{0.3\textwidth}
    \hfill
    \vspace{-4mm}
    \includegraphics[width=0.9\linewidth]{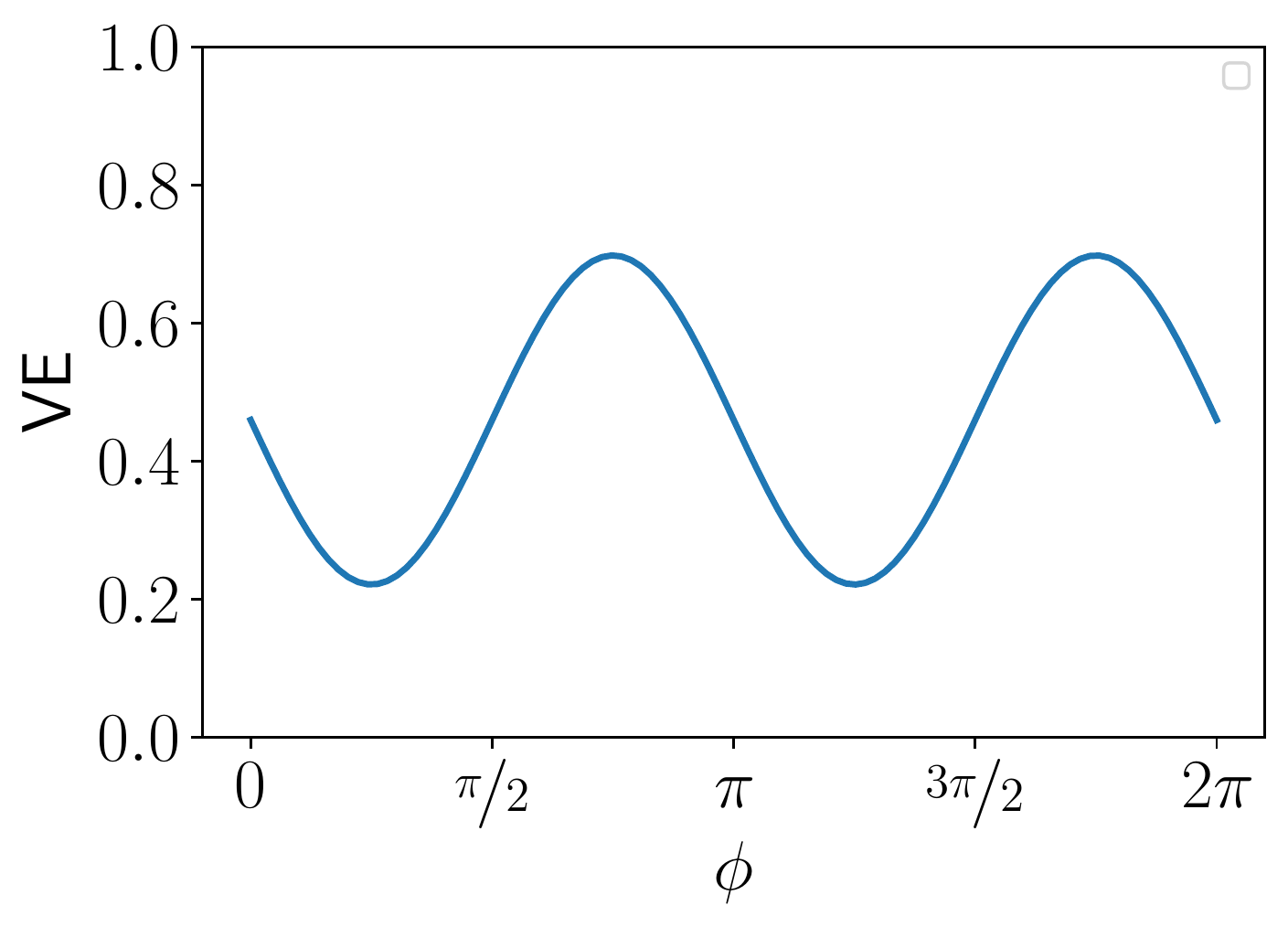}
    \end{subfigure}
    \vspace{-0.2cm}
    \caption{
    \small
    \textbf{Minimizing unconditional MI for the Gaussian linear regression task.}
    To enforce unconditional independence, we choose $\boldW$ such that $\text{Cov}(\boldz)$ is diagonal.
    In our case this is easy: the principal components of $\boldx$ are $x_1 + x_2$ and $x_1 - x_2$.
    The optimal regression loss with minimal MI is then given by whitening and rotating the result by angle $\phi_{\text{opt}}$ which leads to maximal variance explained ($\phi_{\text{opt}} = - \nicefrac{\pi}{4}$ for positive correlations and $\boldA=\boldI$).}
    \label{fig:linear_toy_unconditional}
\end{figure}

\subsection{Unconditional Disentanglement Fails Under Correlation Shift}
\label{sec:unconditional-disentanglement}
In the 2D linear case, we have:
\begin{equation}
\boldz = (z_1, z_2) = \boldW \boldx , \qquad    \widehat{s}_1 = R_1 z_1 , \qquad \widehat{s}_2 = R_2 z_2
\end{equation}
where the matrix $\boldW$ encodes the observation into the latent space.
The linear regression example in Sec.~\ref{sec:supervised-disentanglement} corresponds to $\boldW=\mathbf{C}_{\mathbf{sx}}\mathbf{C}_{\mathbf{x}}^{-1}$ and $R_k = 1$.
In standard supervised objectives, there is no constraint preventing a subspace $z_k$ from containing information about other source variables than $s_k$.
A common approach to enforce independence is to minimize the MI between the latent subspaces $z_1$ and $z_2$~\citep{chen2018isolating,peng2019domain}.
In the Gaussian case, random variables are independent if and only if they are \textit{uncorrelated}.
The optimal linear regression weights $\boldW$ that yield $I(z_1 ; z_2) = 0$ (e.g., such that $\text{Cov}(\boldz)$ is diagonal) can be computed by whitening $\boldx$ and rotating the result by an angle $\phi_{\text{opt}}$ which leads to maximal variance explained.
For our example in Table~\ref{table:regression}, where we have positive correlation and $\boldA = \boldI$, the optimal rotation is $\phi_{\text{opt}} = - \nicefrac{\pi}{4}$ (see Figure~\ref{fig:linear_toy_unconditional}).
However, the resulting model no longer performs well on in-distribution data (Table~\ref{table:regression}, column \wMI).
There is correlation between the source variables $s_1$ and $s_2$ and therefore $I(s_1; s_2) > 0$.
By enforcing independence, at least one of the subspaces cannot contain all relevant information about its attribute and thus will have poor predictive performance.
We make this precise in the following proposition.
\begin{proposition} \label{prop:mi}
If $\text{I}(s_1; s_2) > 0$, then enforcing $\text{I}(z_1; z_2)=0$ leads to $\text{I}(z_k; s_k) < H(s_k)$ for at least one $k$.
\end{proposition}
\vspace{-\baselineskip}
\begin{proof}
The proof is provided in Appendix~\ref{app:proofs}.
\end{proof}

\vspace{-0.1cm}
\subsection{Conditional Disentanglement is Robust to Correlation Shift}
\label{sec:cmi-is-robust}
\vspace{-0.1cm}
\begin{wrapfigure}[20]{r}{0.4\linewidth}
    \vspace{-0.4cm}
    \centering
    \begin{subfigure}{0.33\linewidth}
    \includegraphics[width=\linewidth]{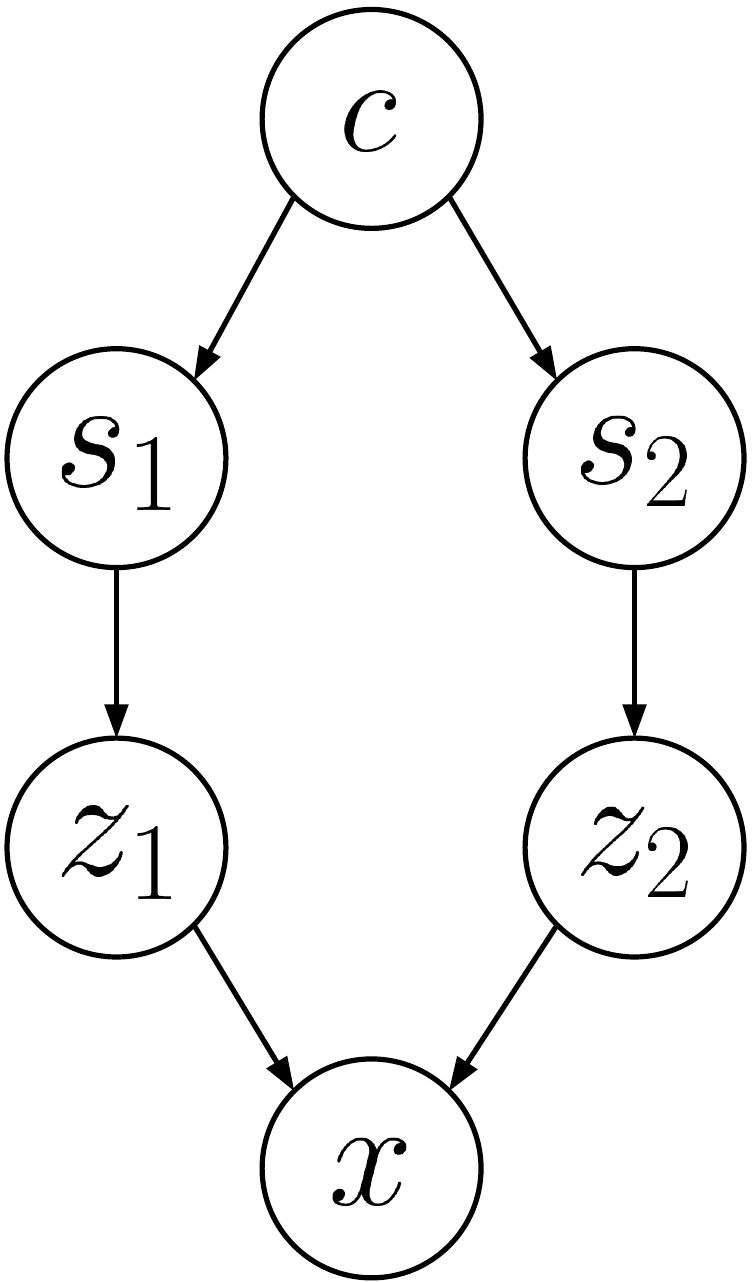}
    \caption*{\footnotesize $I(z_1; z_2) > 0$}
    \end{subfigure}
    \qquad
    \begin{subfigure}{0.33\linewidth}
    \includegraphics[width=\linewidth]{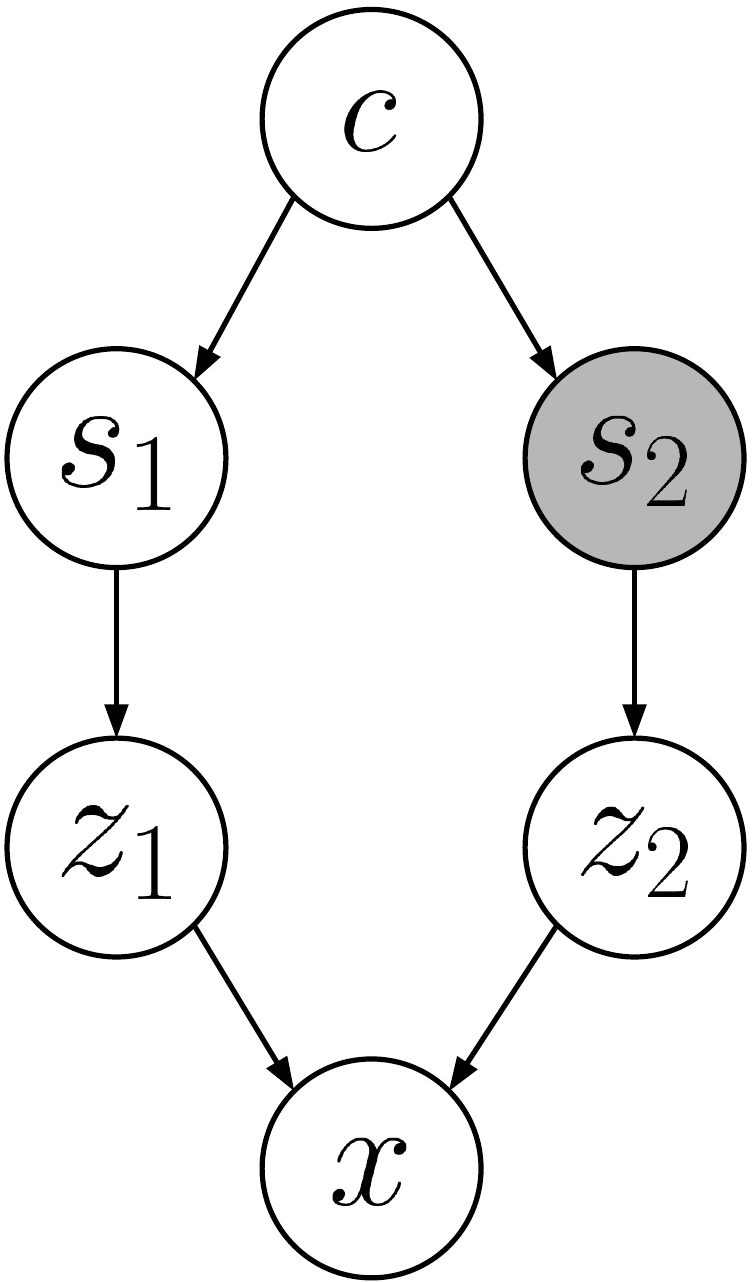}
    \caption*{\footnotesize $I(z_1; z_2 | s_2) = 0$}
    \end{subfigure}
    \caption{
    \small
    The graphical model for two sources $s_1, s_2$ and corresponding latent subspaces $z_1, z_2$. We assume the source variables have a common cause $c$.
    In (a), when none of the sources are observed,  there is a path from $z_1$ to $z_2$, so we have $I(z_1 ; z_2) > 0$; in (b) we observe $s_2$, which breaks the path, and thus $I(z_1 ; z_2 \mid s_2) = 0$.
    }
    \label{fig:graphical-model}
\end{wrapfigure}
We have seen that enforcing unconditional independence between the latent spaces does not solve the disentanglement problem.
% (Appendix~\ref{app:exp-details}, Fig.~\ref{fig:graphical-model})
However, considering the graphical model in Figure~\ref{fig:graphical-model}, $\boldz_1$ and $\boldz_2$ are independent \textit{conditioned on either of $s_1$ or $s_2$}:
assuming a common cause for the correlation between $s_1$ and $s_2$, there is a connection in the graphical model between $\boldz_1$ and $\boldz_2$ introducing a statistical dependence. Observing either $s_1$ or $s_2$ disconnects $\boldz_1$ and $\boldz_2$.
Here, we show that enforcing independence \textit{conditioned on each of the source variables} is also sufficient to yield a robust disentangled representation.
For our 2D example, enforcing conditional independence corresponds to:
\begin{align} 
\text{I}(\boldz_1; \boldz_2 \mid s_1)=0 \qquad \text{and} \qquad \text{I}(\boldz_1; \boldz_2 \mid s_2)=0
\label{eq:cond-independence}
\end{align}
Intuitively, if $s_1$ and $s_2$ are correlated, then $I(s_1 ; s_2) > 0$ and knowing $s_1$ gives us information about $s_2$.
If we can predict $s_1$ from $\boldz_1$, and $s_1$ tells us about $s_2$, then it must be the case that $\boldz_1$ contains information about $s_2$.

We wish to ensure that $\boldz_1$ and $\boldz_2$ share \textit{as little information as possible} (given the ground-truth correlation), to improve robustness to shifts.
Since $\boldz_1$ necessarily contains some information about $s_2$, we enforce that it does not contain \textit{any more information about $\boldz_2$ than necessary} via $I(\boldz_1 ; \boldz_2 | s_2)$, which states that if we know $s_2$, then knowing $\boldz_1$ does not give us more information about $\boldz_2$.

This does not penalize $\boldz_1$ for containing information about $s_2$ due to correctly predicting the correlated variable $s_1$ (and vice versa).
In contrast to MI, this removes only the shared information which is not robust under correlation shift, but keeps the shared information which is necessary to account for the correlation between the source variables.
The optimal solution under the conditional independence constraint (Eq.~\ref{eq:cond-independence}) is achieved by the mapping $\boldW = \boldA^{-1}$, successfully recovering the underlying generative model.
This demonstrates the usefulness of minimizing CMI for generalization under correlation shifts in the case of linear regression with Gaussian variables and motivates us to investigate CMI minimization for larger-scale tasks.

\vspace{-0.2cm}
\section{Method: Minimizing CMI}
\vspace{-0.2cm}
\label{sec:method}
For simple cases such as linear regression, we can compute and minimize the MI and CMI analytically; however, for most tasks, there is no closed form for the mutual information.
In this section, we describe an approach to minimize the CMI for general classification tasks.
Suppose we have a dataset $\mathcal{D} = \{ (\boldx^{(i)}, \bolds^{(i)}) \}_{i=1}^N$ where $\boldx^{(i)}$ is an example and $\bolds^{(i)}$ is a vector of attribute labels --- $\bolds_k^{(i)}$ is the label for the $k^{\text{th}}$ attribute of the $i^{\text{th}}$ example.
We consider discrete attributes, $\bolds^{(i)}_k\in \mathbb{N}$.
Let $f_{\boldtheta} : \boldx \mapsto \boldz$ denote an encoder parameterized by $\boldtheta$ that maps examples $\boldx \in \mathbb{R}^m$ to latent representations $\boldz \in \mathbb{R}^n$.
We aim to learn one latent subspace per attribute, such that each subspace is independent from all other subspaces conditioned on the attribute it encodes.
\begin{figure*}[t]
    \centering
    \includegraphics[width=0.75\linewidth]{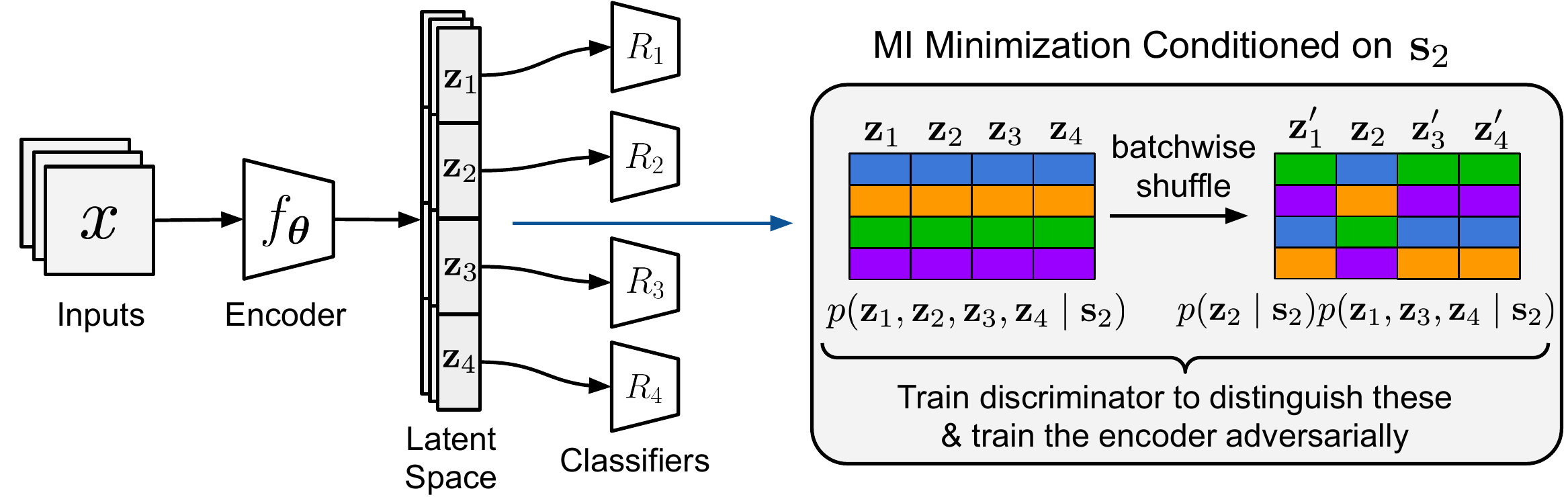}
    \caption{
    \small
    \textbf{Adversarial minimization of conditional mutual information via latent-space shuffling.}
    We minimize the CMI between latent subspaces, $I(\boldz_1 ; \cdots ; \boldz_K \mid \bolds_k)$.
    Here, we illustrate the algorithm for four attributes with corresponding latent spaces $\{ \boldz_1, \boldz_2, \boldz_3, \boldz_4 \}$, where we condition on attribute $\bolds_2$.
    See Sec.~\ref{sec:method} for a description of the method.
    }
    \vspace{-0.1cm}
    \label{fig:method-diagram}
\end{figure*}

\begin{algorithm*}[t]
	\caption{Adversarial Learning of  Conditionally Disentangled Subspaces --- Training the Encoder}
	\label{alg:disentanglement-cond-mi}
	\begin{algorithmic}[1]
	    \State \textbf{Input:} $\{ \boldphi_1, \dots, \boldphi_K \}$, initial parameters for $K$ linear classifiers $R_1, \dots, R_K$
	    \State \textbf{Input:} $\boldtheta$, initial parameters for the encoder $f$
	    \State \textbf{Input:} $\alpha, \beta$ learning rates for training the encoder and linear classifiers
	    \While {true}
	        \State $(\boldx, \{\bolds_k\}_{k=1}^K) \sim \mathcal{D}_{\text{Train}}$  \algorithmiccomment{Sample a minibatch of data with attribute labels}
	        \State $\boldz \gets f_{\boldtheta}(\boldx)$  \algorithmiccomment{Forward pass through the encoder}
        	\State $\{ \boldz_k \}_{k=1}^K \gets \text{SplitSubspaces}(\boldz, K)$  \algorithmiccomment{Partition the latent space into $K$ subspaces}
	        \State $L \gets \sum_{k=1}^K L_{\text{cls}}(R_k(\boldz_k ; \boldphi_k), \bolds_k)$ \algorithmiccomment{Cross-entropy for each attribute}
	        \For{$k \in \{ 1, \dots, K \}$}  \algorithmiccomment{For each attribute/subspace}
	            \State $\boldz' \sim p(\boldz_1, \dots \boldz_K \mid \bolds_k)$ \algorithmiccomment{Samples from the joint distribution}
	            \State $\boldz'' \sim p(\boldz_k \mid \bolds_k) p(\boldz_{-k} \mid \bolds_k)$ \algorithmiccomment{Samples w/ batchwise-shuffled subspaces}
	           \State $L \gets L + \log \left( 1 - D_{\boldomega}(\boldz'') \right) + \log \left( D_{\boldomega}(\boldz') \right)$ \algorithmiccomment{Add adversarial loss}
	        \EndFor
	       \State $\boldtheta \gets \boldtheta - \alpha \nabla_{\boldtheta} L$  \algorithmiccomment{Update encoder parameters}
	       \State $\boldphi_k \gets \boldphi_k - \beta \nabla_{\boldphi_k} L \quad , \quad \forall k \in \{1, \dots, K \}$  \algorithmiccomment{Update classifier parameters}
	   \EndWhile
	\end{algorithmic}
\end{algorithm*}
We have $I(\boldx ; \boldy \mid \boldz) = 0$ if $p(\boldx, \boldy \mid \boldz) = p(\boldx \mid \boldz)p(\boldy \mid \boldz)$.
Our method enforces the latter condition using an adversarial discriminator.
\newcommand{\boldzk}{\mathbf{z}_k}
To obtain samples from $p(\boldz_1, \dots, \mathbf{z}_K \mid \bolds_k)$ and $ p(\boldz_k \mid \bolds_k)p(\boldz_{-k} \mid \bolds_k)$,
we loop over values of $\bolds_k$, and for each condition $\{ \bolds_k = 0, \bolds_k = 1, \dots\}$, we select examples from the minibatch that satisfy the condition, giving us samples from $p(\boldz_1, \dots, \mathbf{z}_K \mid \bolds_k)$; then we shuffle the latent subspaces $\boldz_j, \forall j \neq k$ jointly batchwise (e.g., combining $\boldz_k$ from one example with $\boldz_{-k}$ from another) to obtain samples from $p(\boldz_k \mid \bolds_k) p(\boldz_{-k} \mid \bolds_k)$.
To enforce $p(\boldz_1, \dots, \mathbf{z}_K \mid \bolds_k) = p(\boldz_k \mid \bolds_k)p(\boldz_{-k} \mid \bolds_k)$, we train the encoder $f$ adversarially against a discriminator trained to distinguish between these two distributions.
The discriminator takes as input a representation and predicts whether it is ``real'' (e.g., drawn from the joint distribution) or ``fake'' (e.g., drawn from the product of marginals).
One discriminator is trained for each attribute $\bolds_k$, which receives samples from the two distributions and the attribute value it is conditioned on.
In practice, we use a conditional discriminator, effectively sharing parameters between the discriminators for each of the attributes.
This process is illustrated in Figure~\ref{fig:method-diagram}.
Algorithm~\ref{alg:disentanglement-cond-mi} describes the encoder training loop; Algorithm~\ref{alg:disentanglement-cond-mi-train-disc} in Appendix~\ref{app:algorithms} describes the corresponding discriminator training loop.
We formally describe the algorithms for the baselines (\oBase and \textit{Base + MI}) in Appendix~\ref{app:algorithms}.

This approach is architecture-agnostic, and can be used to factorize the latent space of any classifier or generative model (e.g., VAEs~\citep{joy2020capturing} or flow-based models~\citep{kingma2018glow}).
However, some models (such as VAEs) may have objectives that interfere with the goal of obtaining conditionally independent subspaces; for example, the ELBO encourages independence between all latent dimensions.
In our experiments, we used linear and MLP encoders rather than VAEs to avoid this conflicting objective.

Because the latent space is typically low-dimensional, we have a choice of different distribution alignment techniques, including maximum mean discrepancy (MMD)~\citep{gretton2006kernel} and adversarial approaches~\citep{goodfellow2014generative}.
Different GAN formulations can be interpreted as minimizing different divergences: the vanilla GAN~\citep{goodfellow2014generative} minimizes the Jensen-Shannon divergence; WGAN~\citep{arjovsky2017wasserstein} minimizes the Wasserstein distance, which has been used to define an analogue of mutual information called the \textit{Wasserstein dependency measure}~\citep{ozair2019wasserstein}; $f$-GAN~\citep{nowozin2016f} minimizes an arbitrary $f$-divergence, etc.
Each of these divergence measures will be 0 if and only if the subspaces are independent, however their training dynamics may differ.
In practice, we found the vanilla GAN formulation to work well across our experiments.

\vspace{-0.1cm}
\section{Experiments}
\label{sec:experiments}
\vspace{-0.1cm}

\begin{figure*}[t]
\centering
\begin{subfigure}{0.46\linewidth}
\includegraphics[width=\linewidth]{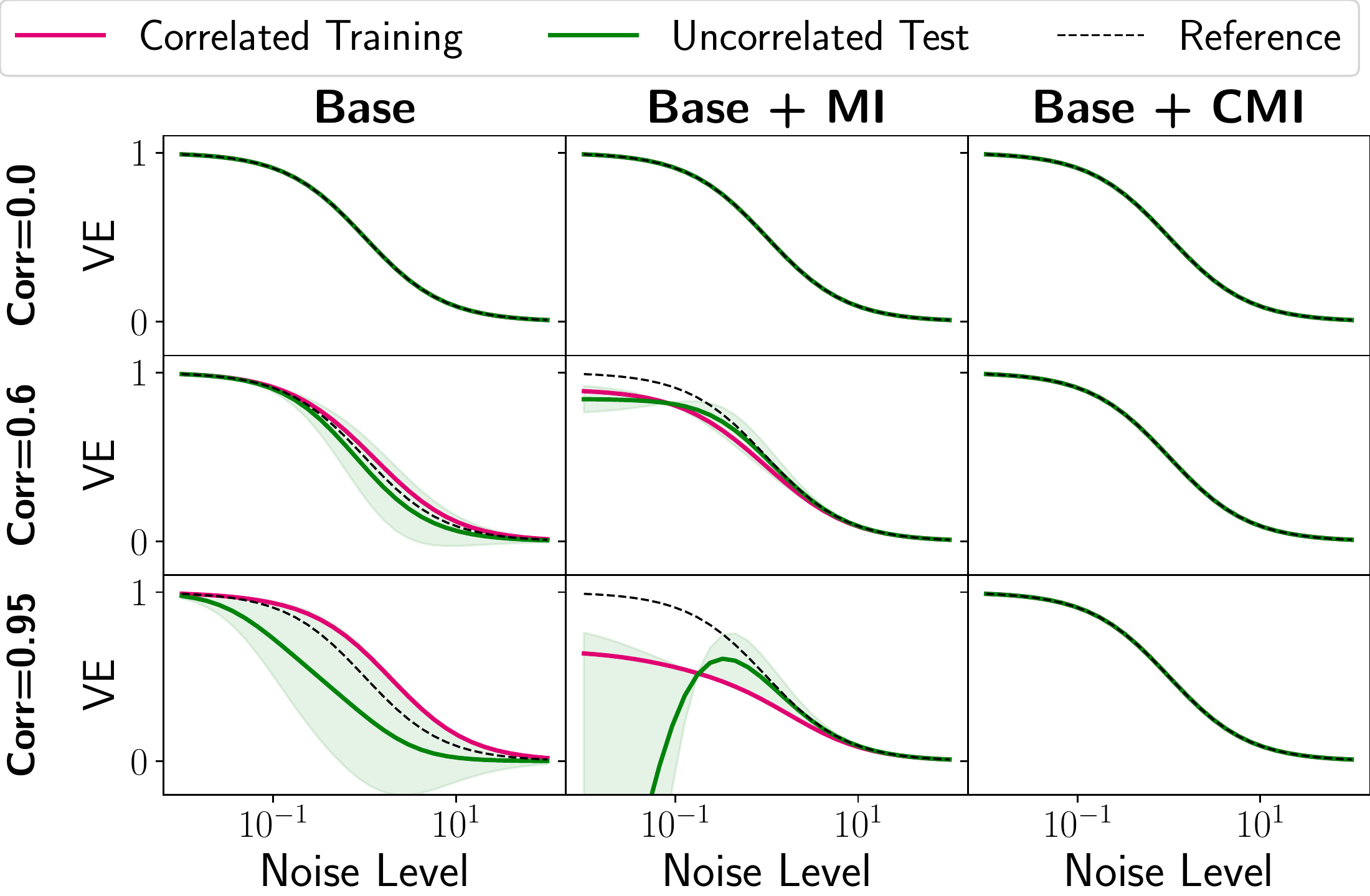}
\caption{Toy linear regression.}
\label{fig:regression_vary}
\end{subfigure}
\qquad
\begin{subfigure}{0.46\linewidth}
\includegraphics[width=\linewidth]{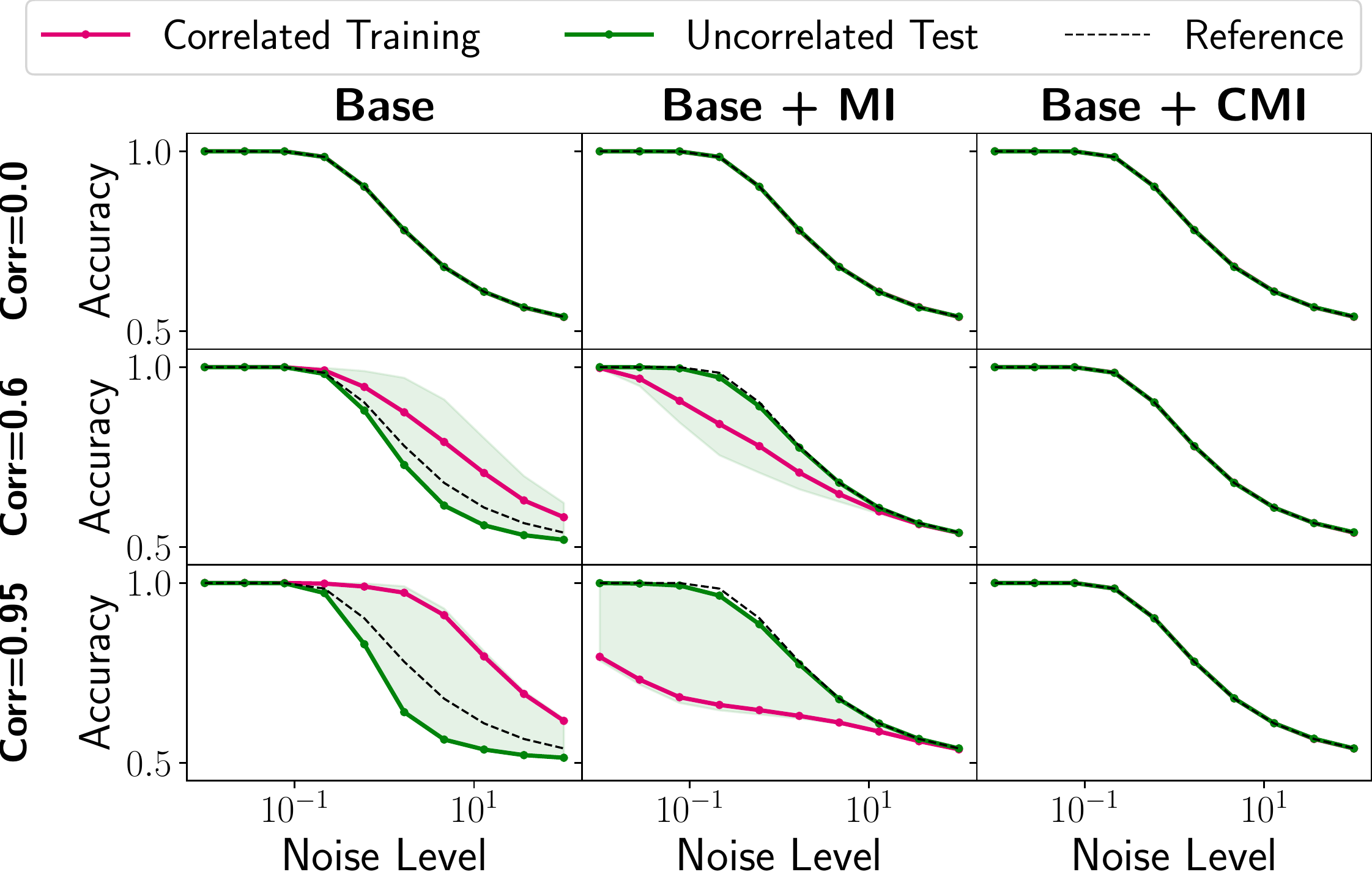}
\caption{Toy classification with ten attributes.}
\label{fig:classification}
\end{subfigure}
\vspace{-0.2cm}
\caption{
\small
\textbf{Synthetic linear regression (left) and linear classification (right) tasks.}
We measure the performance (variance explained for regression and accuracy for classification) on the correlated training data ({\color{magenta}magenta}) and on test data with a range of correlation shifts ({\color{dkgreen}green}, solid line is the uncorrelated test data). The performance of the \oBase model in the uncorrelated setting serves as a reference in each plot (dashed black line) and facilitates the comparison of the performance of the different objectives (columns). In both tasks, we find that, \oCMI leads to robustness to correlation shift independent of the noise level (x-axis) and the strength of the correlation in the training data (rows), while the other approaches do not.
}
\vspace{-0.4cm}
\label{fig:toy-regression-cls}
\end{figure*}

Our experiments aim to answer the following questions: 1) What is the effect of the train-time correlation strength and noise level on the solutions found by training with each objective, \oBase, \oMI, and \oCMI? 2) Can we successfully learn conditionally disentangled representations for classification tasks using Algorithm~\ref{alg:disentanglement-cond-mi}? and 3) Does CMI minimization lead to improved correlation-shift robustness on natural image datasets including MNIST and CelebA?

First, we present results on the analytically-solvable linear regression example, illustrating the effect of the correlation strength and noise level on the solution obtained by each objective.
Then, we demonstrate that our findings also hold for a synthetic classification task with multiple attributes.
Next, we employ the method described in Section~\ref{sec:method} and investigate two realistic tasks, a multi-digit MNIST task with occlusions and correlated CelebA, and show that minimizing CMI can largely eliminate the gap in performance caused by test-time correlation shifts.
Finally, we evaluate common disentanglement metrics and apply Algorithm~\ref{alg:disentanglement-cond-mi} in weakly supervised settings.
Experimental details and extended results are provided in Appendix~\ref{app:exp-details}.

\paragraph{Linear Regression.}

\label{sec:toy-linear-reg}
Here, we revisit the linear regression problem from Section~\ref{sec:linear-reg}, to investigate the impact of the train-time correlation strength and noise level on the models learned with each of the objectives \oBase, \oMI, and \oCMI.
The results are shown in Figure~\ref{fig:regression_vary}.
We found that \oCMI yields robustness to correlation shift across all correlation strengths and noise levels, while the baselines do not.
The performance of \oBase drops most severely under correlation shift for strong train-time correlations and intermediate noise levels; in this regime, \oCMI improves performance substantially.

\vspace{-0.1cm}
\paragraph{Toy Multi-Attribute Classification.}
\label{sec:toy-cls}

Next, we investigated whether these findings hold for classification tasks with multiple attributes.
Here, binary source attributes $s_k = \pm 1$, $\forall k \in \{ 1, \dots, K \}$ generate the observed data via $\mathbf{x} = \boldA \mathbf{s} + \mathbf{n}$ (we set $\boldA=\boldI$ for simplicity) with normally distributed noise $\mathbf{n}\sim \mathcal{N}(\mathbf{0}, \mathbf{C_n})$.
We induced correlations between the attributes $a_k$, such that the number of datapoints differs for the different combinations of attribute values.
In the multi-attribute setting, the correlation strength refers to the pairwise correlation between all attributes.
Similarly to the regression task, we find that \oCMI leads to robustness under correlation shift (see Figure~\ref{fig:classification} and Appendix~\ref{app:toy-multi-attribute-cls}).

\begin{figure*}[t]
\centering
\vspace{-0.2cm}
    \begin{subfigure}{0.56\textwidth}
      \includegraphics[width=\linewidth]{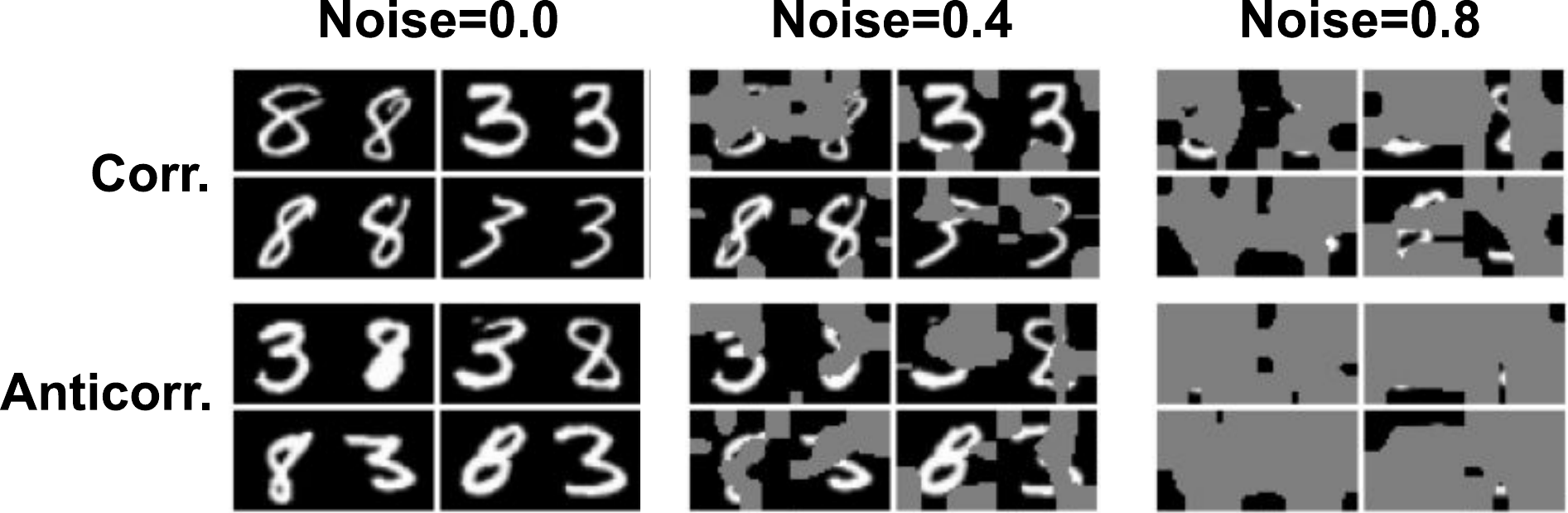}
      \caption{Correlated training data and anti-correlated test data.}
      \label{fig:multi-mnist-occlusions}
    \end{subfigure}
    \hfill
    \begin{subfigure}{0.4\textwidth}
        \includegraphics[width=\linewidth]{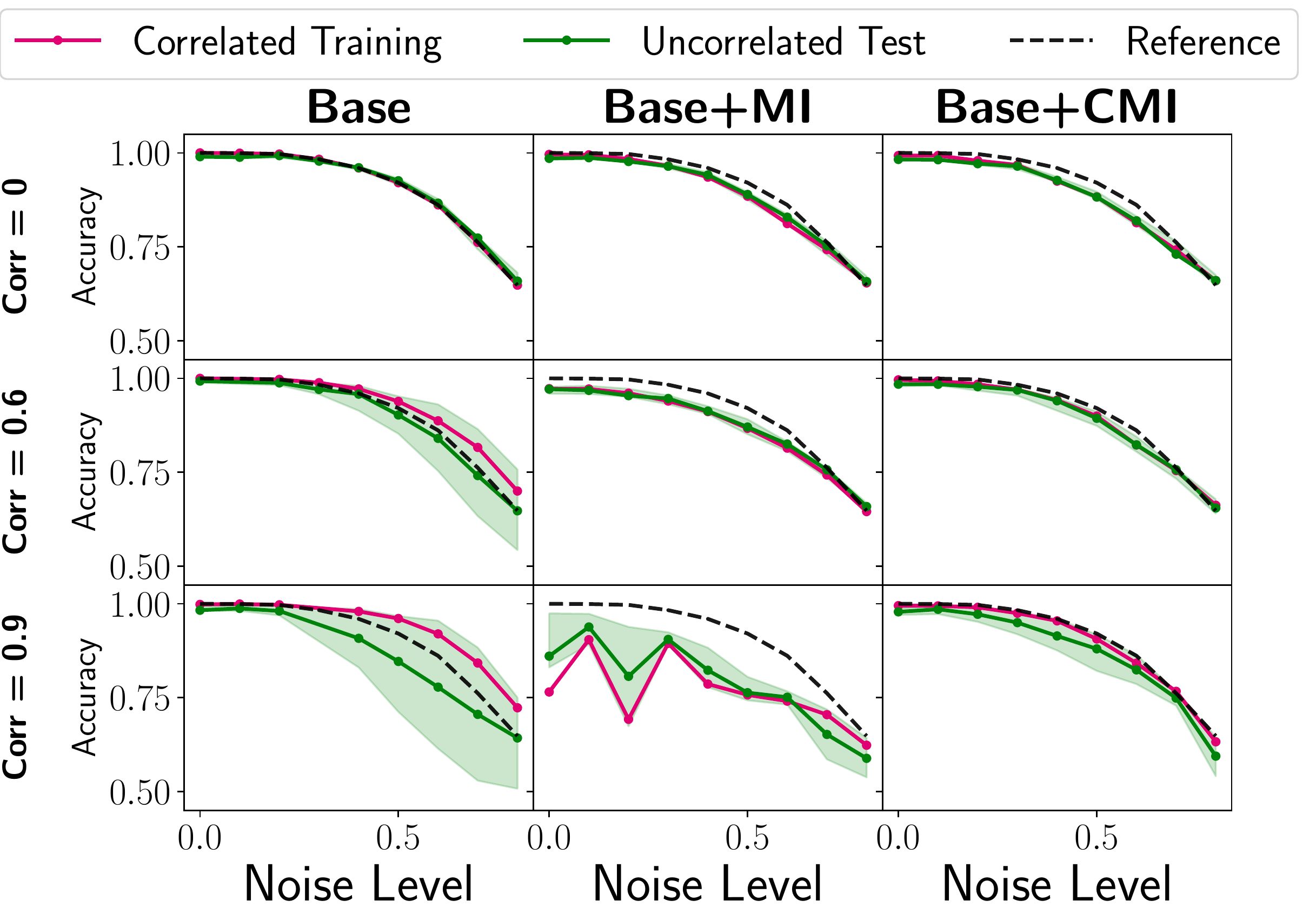}
        \caption{Performance comparison.}
        \label{fig:mnist-results}
    \end{subfigure}
    \vspace{-0.1cm}
    \caption{\small \textbf{Multi-digit occluded MNIST.} (a) Examples of the correlated training data (where \texttt{3-3} and \texttt{8-8} pairs are frequent) and anticorrelated test data (where \texttt{3-8} and \texttt{8-3} pairs are frequent), under a range of occlusion strengths. (b) Accuracies under correlation shifts for different noise levels, achieved by training with each of the objective functions \oBase, \oMI, and \oCMI.
    \oCMI achieves consistent performance across correlation shifts.
    Similarly to Figure~\ref{fig:toy-regression-cls}, here we show the reference performance of the model trained on uncorrelated data (solid black line), the performance on correlated training data ({\color{magenta}magenta}) and on a range of test-time correlations in $[0,1]$ (shaded {\color{dkgreen}green} region, where solid green denotes the uncorrelated test performance).
    }
    \label{fig:multi-mnist}
    \vspace{-0.4cm}
\end{figure*}

\vspace{-0.1cm}
\paragraph{Multi-Digit Occluded MNIST.}
\label{sec:mnist}

Next, we designed a larger-scale task to investigate whether these properties hold in a more complex setting.
We created a dataset by concatenating two MNIST digits side-by-side, where the aim is to predict both the left- and right-hand labels.
We generated occlusion masks using the procedure used by~\cite{chai2021using}; examples from our synthetic dataset under a range of noise settings are shown in Figure~\ref{fig:multi-mnist-occlusions}.
We used a subset of MNIST consisting of classes \texttt{3} and \texttt{8} (which are visually similar and can become ambiguous under occlusions).
This mimics multiple-object classification in a way that allows us to control the correlation strength and noise level (via the amount of occlusion), allowing for systematic analysis.
This task is a more complex analogue of the synthetic classification task from Figure~\ref{fig:classification}.
We added explicit occlusion noise because the MNIST data itself is simple, and has too little ``natural'' noise to clearly observe the predicted effects (e.g., for low noise levels, the supervised loss already does well).
While this task would also be possible for colored MNIST and dSprites, one advantage of our task is its symmetry, which allows us to exclude potential side-effects: here, the attributes have the same type (the digit identity), whereas the attributes in colored MNIST (digit identity and color) and dSprites (shape, size, position, etc.) are more diverse.

Similarly to the toy tasks, we train an encoder to map images onto a $D$-dimensional latent space, which is partitioned in two equal-sized subspaces corresponding to the two digits; we train a linear classifier on each subspace to predict the respective class labels.
We consider different correlation strengths between the left and right digits in the training set (where strong correlation means that the digits often match, e.g., \texttt{3-3} or \texttt{8-8} are more common than \texttt{3-8} or \texttt{8-3}).
We evaluate each model on test data with correlation strengths ranging from $[-1, 1]$.
The results are shown in Figure~\ref{fig:mnist-results}.
We found that the conclusions from the toy experiments hold in this setting: supervised learning with only the cross-entropy loss, as well as with unconditional MI minimization, fail under test-time correlation shift, while minimizing CMI is more robust.
Experimental details and extended results are provided in Appendix~\ref{app:mnist}.

\begin{figure}[t]
    \centering
    \begin{subfigure}{.25\linewidth}
      \includegraphics[width=\linewidth]{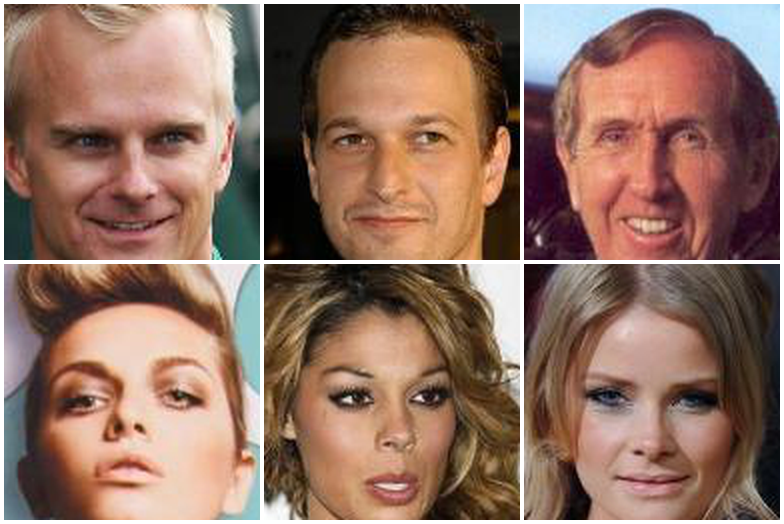}
      \caption{Correlated train data.}
      \label{fig:celeba-train-examples}
    \end{subfigure}
    \begin{subfigure}{.25\linewidth}
        \includegraphics[width=\linewidth]{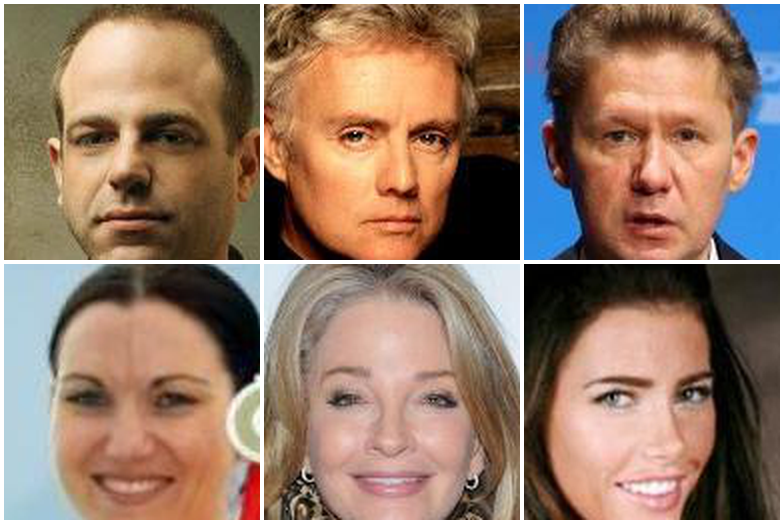}
        \caption{Anti-correlated test data.}
      \label{fig:celeba-test-examples}
    \end{subfigure}%
    \hfill
    \begin{subfigure}{0.48\linewidth}
        \includegraphics[width=\linewidth]{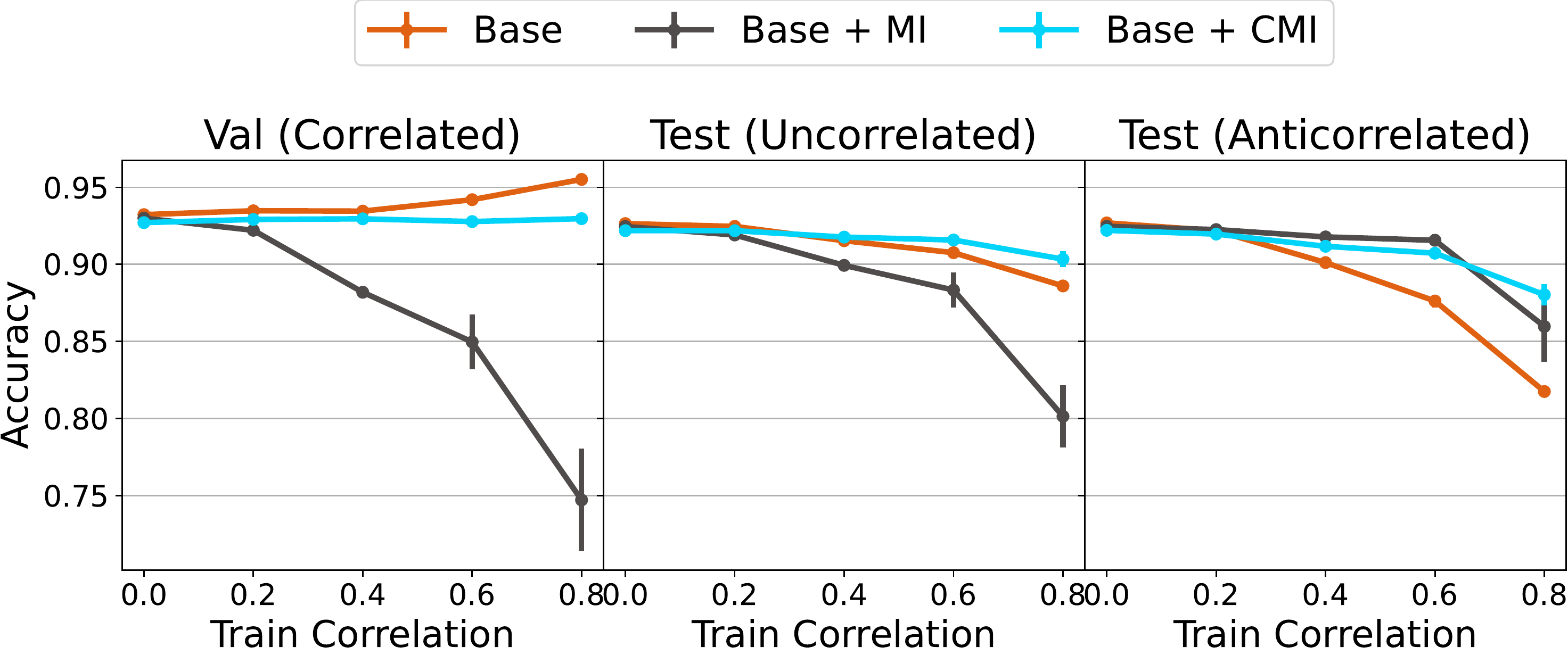}
        \caption{Performance comparison.}
        \label{fig:celeba-accuracies}
    \end{subfigure}
    \vspace{-0.3cm}
    \caption{
    \small
    \textbf{Correlated CelebA.} (a) Training examples with correlation 0.8 between attributes \texttt{Male} and \texttt{Smiling}, such that the majority of men are smiling while the majority of women are not. (b) Anti-correlated test examples, where the majority of women are smiling. (c) Accuracies of each method under a range of correlation strengths, for validation data with the same correlation as the training data, uncorrelated test data, and anticorrelated test data.}
    \label{fig:celeba-corr-acc}
    \vspace{-0.4cm}
\end{figure}

\vspace{-0.1cm}
\paragraph{Correlated CelebA.}
\label{sec:celeba}

Finally, we consider a realistic setting using the CelebA faces dataset~\citep{liu2015faceattributes}.
In contrast to the multi-digit MNIST task, here we do not add any artificial observation noise (as CelebA is a more complex dataset that naturally has noise in observations and/or labels).
We selected two attributes that we know \textit{a priori} are not causally related, \texttt{Male} and \texttt{Smiling}, and we created subsampled datasets with a range of training correlations $\{0, 0.2, 0.4, 0.6, 0.8 \}$.
We evaluated our models on both \textit{anti-correlated} and \textit{uncorrelated} test sets (Figures~\ref{fig:celeba-train-examples} and \ref{fig:celeba-test-examples}).
Figure~\ref{fig:celeba-accuracies} compares the performance of the baseline classifier, unconditional MI model, and conditionally disentangled model under a range of correlation strengths.
We found that minimizing CMI has a larger effect for medium-to-high correlation; however, CMI minimization does not hurt performance at low correlation strengths.
Note that while the unconditional model appears to have good performance on the anti-correlated test set, its performance is poor on the validation set (that has the same correlation structure as the training set), so this model does not perform well on in-distribution-data.
In contrast, the \oCMI model performs well on both in-distribution data and shifted test distributions. 
Also note that the problem of disentangling correlated attributes does not occur only under correlation shift, but is already present in the source domain where certain attribute combinations will reliably be treated incorrectly. For example, \oBase fails to recognize the rare non-smiling male faces in 49\% of the cases, while \oCMI fails only in 25\% of the cases.
Additional details are in Appendix~\ref{app:celeba}.

\textbf{Disentanglement Metrics.}
\cite{locatello2020sober} showed that common disentanglement metrics are not suitable for the correlated setting.
For this reason, we focused on comparing performance under correlation shift, which we consider more suitable for correlated data: if a model cannot predict a factor of variation well for certain values of another factor, then the model did not successfully disentangle those factors.
\begin{wrapfigure}[14]{r}{0.5\linewidth}
\vspace{-0.3cm}
\centering
\includegraphics[width=\linewidth]{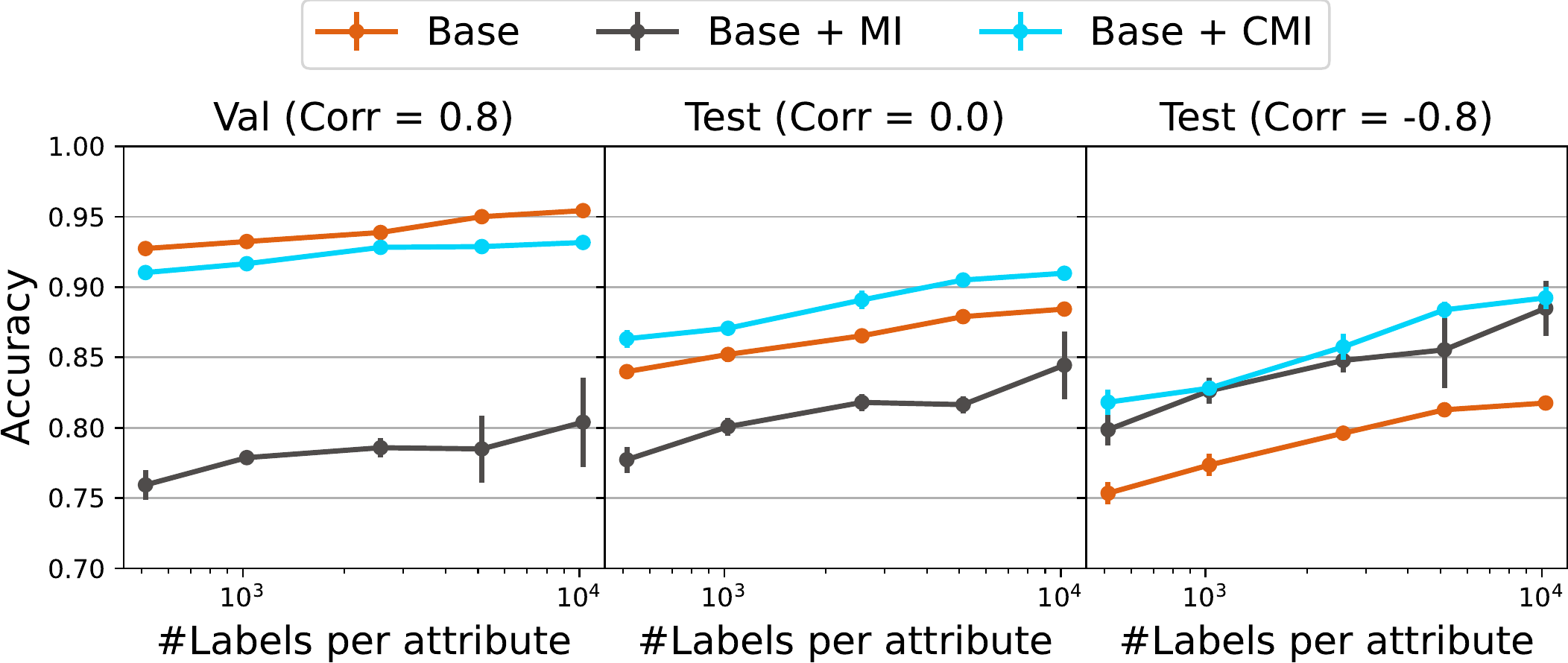}
\caption{
\small
\textbf{Weakly-supervised CelebA.}
The x-axis shows the number of labels per attribute used during training; the rightmost datapoint corresponds to full supervision.
\oCMI outperforms the other objectives under correlation shift.
}
\label{fig:celeba-weakly}
\end{wrapfigure}
However, one can still make use of the disentanglement metrics by evaluating them on \textit{uncorrelated data}, using models trained on correlated data.
We performed this analysis for the toy classification and CelebA tasks, and found that \oCMI leads to improved disentanglement scores across a wide range of metrics, compared to \oBase and \oMI (Appendix \ref{app:dis_metrics}).

\paragraph{Extension to the Weakly Supervised Setting.}
Algorithm~\ref{alg:disentanglement-cond-mi} can be applied directly to weakly supervised settings; it is not necessary for each datapoint to have labels for all attributes.
We find that when reducing the number of labels, \oCMI outperforms the other objectives under correlation shift (see Figure \ref{fig:celeba-weakly} and Appendix~\ref{app:weakly}).

\section{Limitations \& Future Work}
\vspace{-0.3cm}
\label{sec:limitations}
Our study mainly concerns the setting where the underlying factors of variation are known. Practical applications of this setting can occur with respect to fairness, where one may wish to train a model such that correlations that exist in the training data are not relied upon for prediction. Nonetheless, full supervision is a strong assumption and an exciting goal for future work would be to look into relaxing this assumption. Our experiment with the weakly supervised version of the CelebA experiment is a first step in this direction.

We have shown that minimizing CMI yields predictions that disregard correlations between attributes in the training data, which is helpful when correlations shift between the training and test data. This approach relies on knowing a priori which correlations should not be used. This is the case, for example, for fairness applications where a person’s race or gender should not affect the results. A direction for future work would be to automatically determine which correlations are more or less likely to shift in held-out data and to add this step before applying our approach of avoiding the unwanted correlations.
One may incorporate ideas from IRM~\citep{arjovsky2019invariant}, which leverages multiple environments at training time to discover which correlations tend to shift and which are stable---e.g., to distinguish between causal and spurious correlations, the latter of which we wish to avoid relying on.
A fruitful direction for future work would be to combine IRM-style discovery of spurious correlations with our approach, which can be used to control for these correlations when learning disentangled representations. In a related vein, there has been recent work which aims to discover environments when none are given explicitly~\citep{creager2021environment}, which may be useful in combination with our work.

While CMI is defined for both continuous and discrete attributes, our method of shuffling the latent subspaces is only applicable to discrete attributes.
Discrete attributes are prevalent in many settings: in domain adaptation, the class and domain are discrete; in multi-object classification, the class of each object is a discrete attribute; the foreground and background of natural images are discrete, etc. Nevertheless, finding methods to minimize CMI for continuous attributes is an interesting direction for future work. 
Another caveat of our method for minimizing the CMI via latent subspace shuffling is the increased computational cost relative to minimizing the unconditional MI: the cost for CMI scales linearly with the number of attributes and attribute values, while the cost for MI is constant.

\vspace{-0.3cm}
\section{Conclusion}
\vspace{-0.3cm}
\label{sec:conclusion}
Correlations are prevalent in real-world data, yet pose a substantial challenge for disentangled representation learning.
Standard approaches learn to rely on these correlations, especially when data are noisy, as the correlations provide an easy-to-learn signal with predictive power.
When the attributes are not causally related, this leads to poor performance under test-time correlation shift.
Although for small correlations the effects may not be large, relying on these correlations and thereby systematically treating a subset of the data incorrectly, can be catastrophic for fairness.
We first showed that supervised learning and \textit{unconditional} mutual information minimization fail to learn representations robust to such shifts.
We then argued that the correct notion of disentanglement in such cases is \textit{conditional disentanglement}, and we proposed a simple approach to minimize the conditional mutual information between latent subspaces.
We showed that conditionally disentangled representations improve robustness to correlation shift in analytically solvable linear tasks, as well as on natural images.
Overall, we established CMI minimization as a more appropriate alternative to MI minimization, which sets the stage for the development of more powerful objective functions for disentanglement.

\subsubsection*{Acknowledgements}
We thank Jörn-Henrik Jacobsen for his valuable contributions in the early stage of this work.
We thank Steffen Schneider, Dylan Paiton, Lukas Schott, Elliot Creager, and Frederik Träuble for helpful discussions. 
We thank the International Max Planck Research School for Intelligent Systems (IMPRS-IS) for supporting Christina Funke.
Paul Vicol was supported by a JP Morgan AI Fellowship.

We acknowledge support from the German Federal Ministry of Education and Research (BMBF) through the Competence Center for Machine Learning (FKZ 01IS18039A) and the Bernstein Computational Neuroscience Program T\"ubingen (FKZ: 01GQ1002), the German Excellence Initiative through the Centre for Integrative Neuroscience T\"ubingen (EXC307), and the Deutsche Forschungsgemeinschaft (DFG; Projektnummer 276693517 – SFB 1233).
Resources used in preparing this research were provided, in part, by the Province of Ontario, the Government of Canada through CIFAR, and companies sponsoring the Vector Institute www.vectorinstitute.ai/partners.

\bibliography{collas2022_conference}
\bibliographystyle{collas2022_conference}

\clearpage

\onecolumn

\appendix

\section*{Appendix}

This appendix is structured as follows:
\begin{itemize}
    \item In Section~\ref{app:notation} we provide an overview of the notation we use throughout the paper.
    \item In Section~\ref{app:exp-details} we provide experimental details, as well as extended results.
    \item In Section~\ref{app:algorithms} we provide the algorithms for the baseline methods, namely for classification-only training and unconditional mutual information minimization.
    \item In Section~\ref{app:proofs} we provide a proof of Proposition~\ref{prop:mi}.
\end{itemize}

\section{Notation}
\label{app:notation}

\begin{table}[h!]
\begin{center}
    \begin{tabular}{c c}
        \toprule
        \textbf{Symbol} & \textbf{Meaning} \\
        \midrule
        $\boldx$ & Observations \\
        $\bolds$ & Ground-truth latent factors \\
        $\hat{\bolds}$ & Predictions of factors \\
        $\boldz$ & Latent representation \\
        $\boldW$ & Linear regression weights \\
        $R_1, R_2$ & Linear readout from the latent space $\boldz$ to predictions $\hat{\bolds}$ \\
        $\boldn$ & Isotropic Gaussian noise, $\boldn \sim \mathcal{N}(0, \mathbf{C_n})$ with $\mathbf{C_n} = \sigma^2 I$\\ 
        $\boldA$ & Square matrix used to generate observations for the linear task as $\boldx = \boldA \bolds + \boldn$ \\
        $f$ & Encoder function \\
        $f_{\theta}$ & Encoder function with parameters $\theta$ \\
        \bottomrule
    \end{tabular}
\end{center}
\vspace{-0.3cm}
\caption{Summary of the notation used in this paper.}
\label{tab:TableOfNotation}
\end{table}

\section{Experimental Details and Extended Results}
\label{app:exp-details}

\paragraph{Method Details.}
Note that the dimensions $m$ and $n$ are arbitrary---in particular, $n$ does not need to be smaller than $m$.
In principle, each subspace can have different dimension (e.g., the linear readout layer for each attribute can have arbitrary dimensions $A \times S$ where $A$ is the attribute dimensionality and $S$ is the dimensionality of a particular subspace).

\paragraph{Compute Environment.}
Our experiments were implemented using PyTorch~\citep{paszke2019pytorch}, and were run on our internal clusters.
The toy 2D experiments were run on a single NVIDIA RTX 2080 TI GPU, and took approximately 48 hours for all the results presented.
The MNIST and CelebA experiments were run on NVIDIA Titan Xp GPUs.
Each run of the multi-digit MNIST and CelebA tasks for a given method and correlation strength (and noise level in the MNIST case) took approximately 12 hours, and these were run in parallel.

\subsection{Toy Multi-Attribute Classification}
\label{app:toy-multi-attribute-cls}

\begin{wrapfigure}[13]{r}{0.5\linewidth}
\vspace{-.88cm}
\includegraphics[width=\linewidth]{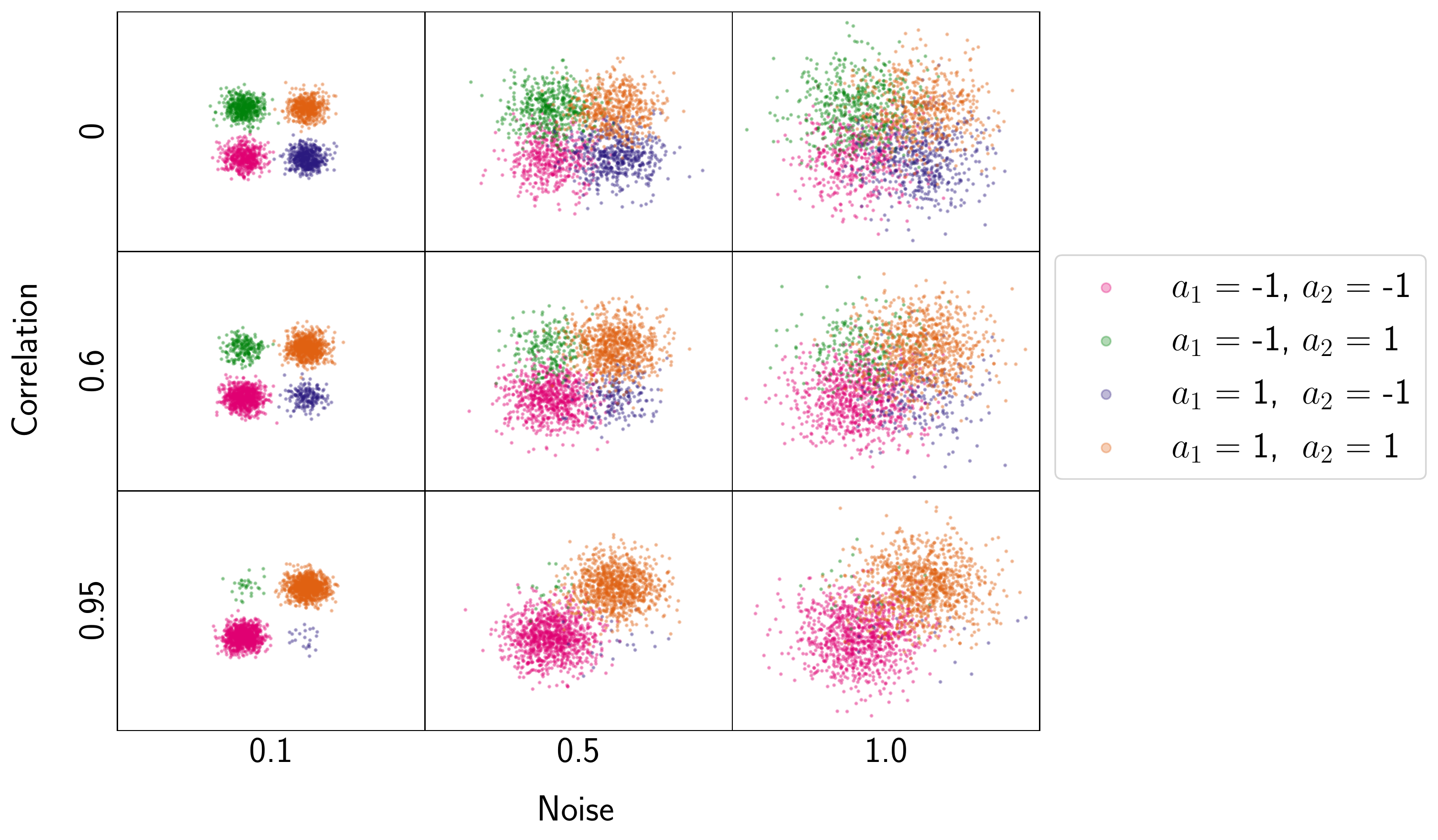}
\vspace{-0.8cm}
\caption{Data used for linear classification with two attributes ($a_1$ and $a_2$), visualized for a range of correlation strengths and noise levels.} 
\label{fig:linear-cls-data}
\end{wrapfigure}
\begin{figure*}
\begin{subfigure}{.32\textwidth}
  \includegraphics[width=1\linewidth]{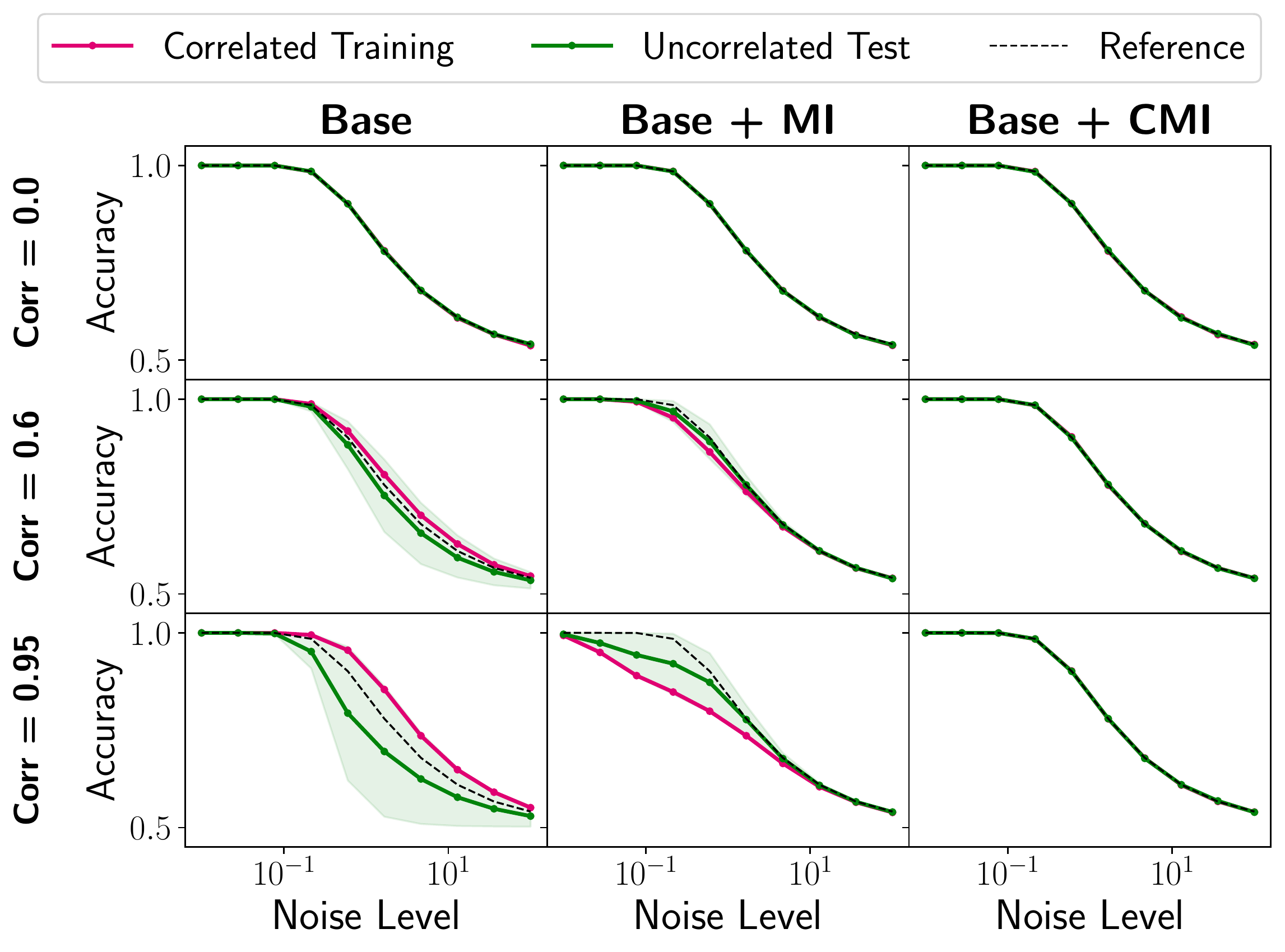}
  \vspace{-0.3cm}
  \caption{Two attributes.}
\end{subfigure}%
\hfill
\begin{subfigure}{.32\textwidth}
  \centering
  \includegraphics[width=1\linewidth]{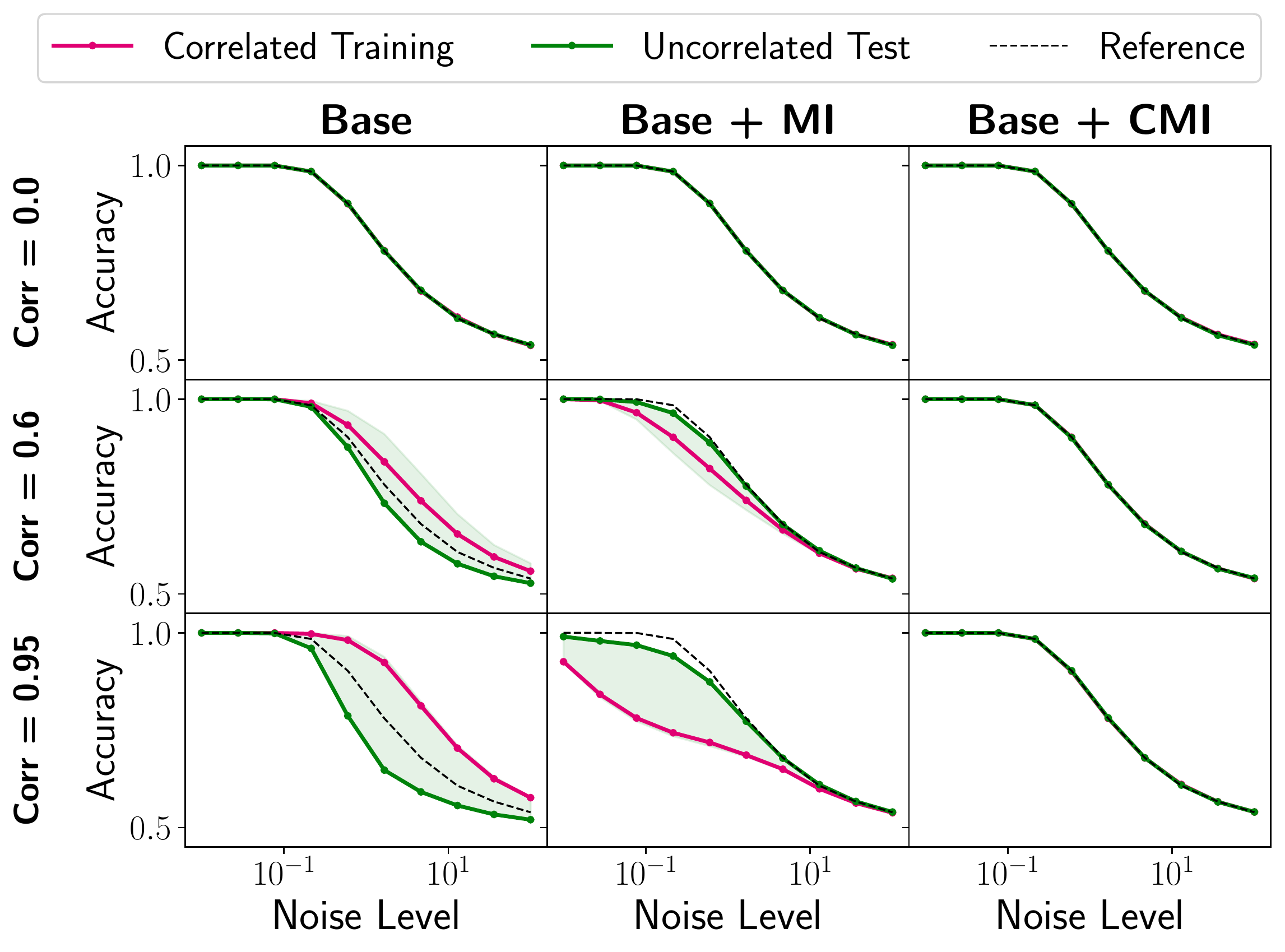}
  \vspace{-0.3cm}
  \caption{Four attributes.}
\end{subfigure}
\hfill
\begin{subfigure}{.32\textwidth}
  \centering
  \includegraphics[width=1\linewidth]{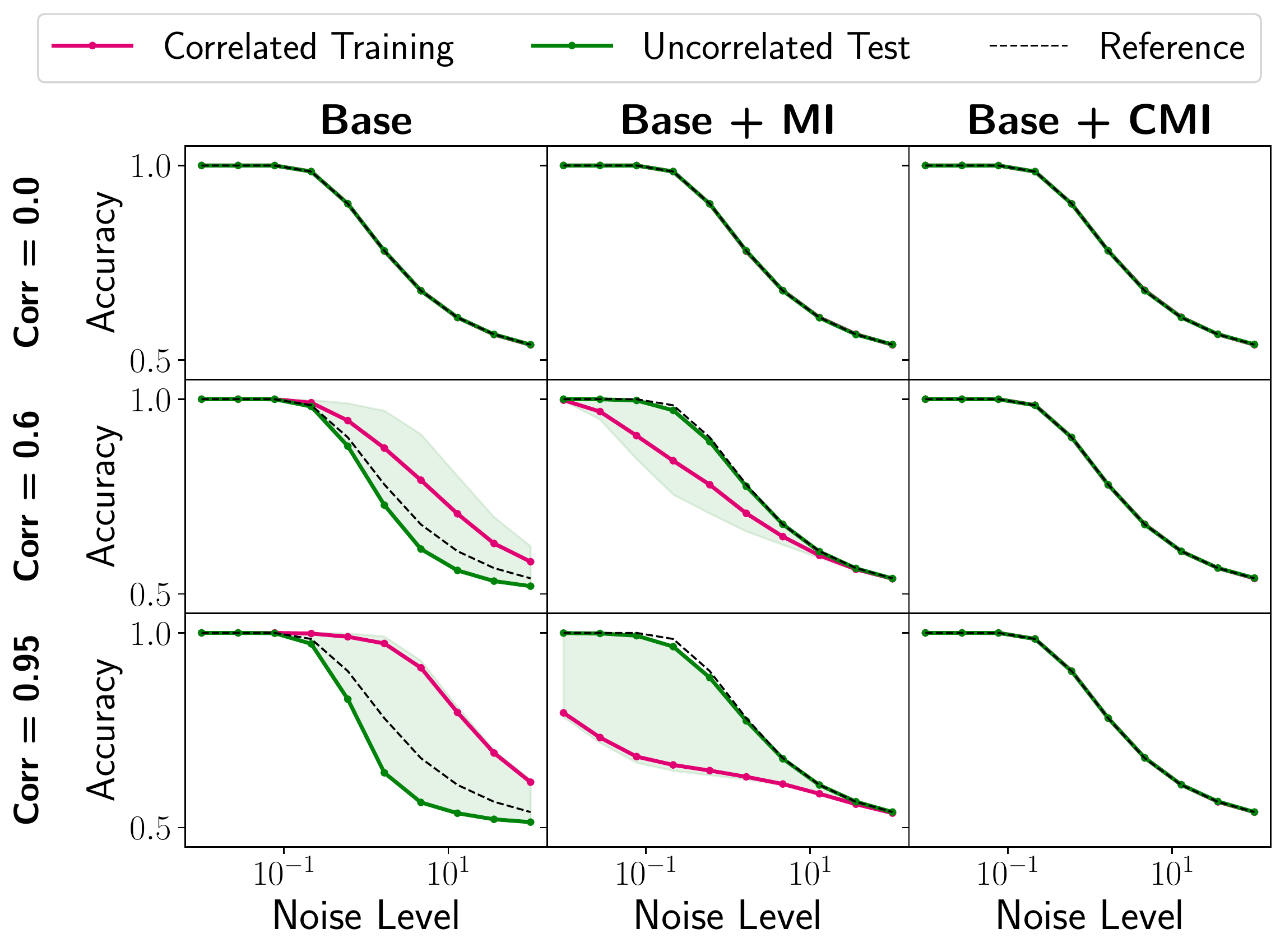}
  \vspace{-0.3cm}
  \caption{Ten attributes.}
\end{subfigure}
\caption{Toy classification with different numbers of attributes. Strong negative correlations could not be generated for multiple attributes; thus only positive test correlations were evaluated for (b) and (c).}
\vspace{-0.3cm}
\label{fig:classification_app}
\end{figure*}

We performed this experiment with two, four and ten binary attributes. The results for varying numbers of attributes are shown in Figure~\ref{fig:classification_app}. For two attributes we illustrated the data $\boldx$ for different correlation strength and noise levels (Figure~\ref{fig:linear-cls-data}). Here, increasing the correlation strength means that data points with $a_1 = a_2$ are increasingly more common relative to $a_1 \neq a_2$. The noise level on the other hand determines the overlap of the distributions and therefore the difficulty of the task. 

\paragraph{Experimental Details.} We used a PacGAN-style setup~\citep{lin2018pacgan} for our toy experiments, where the discriminator takes as input a concatenation of 50 samples. 
\vspace{0.3cm}
\begin{itemize}
    \itemsep0em 
    \item \textbf{Base}: We used Adam~\citep{kingma2014adam} with a learning rate of 0.01.
    \item \textbf{Base + MI}: We used Adam to optimize the encoder, linear classifiers, and discriminators. After each step of optimizing the discriminator and encoder, we optimized the linear classifiers ($R$) for 10 steps. The disentanglement loss term was weighted by a factor of 100 relative to the classification loss. In preliminary tests, we found that the optimal learning rate depended on noise level, correlation strength, and number of attributes. The results in Figure \ref{fig:classification} were obtained using one of the following learning rates for the discriminator $\{1e-4, 2e-4, 5e-4, 1e-3, 5e-3\}$. The learning rate of the generator and linear classifiers was chosen to be 10 times smaller than the discriminator learning rate.
    \item \textbf{Base + CMI}: For $\boldA=\boldI$, no optimization was necessary, as we already know the optimal solution to be $\boldW=\boldA^{-1}=\boldI$. We confirmed experimentally that the discriminator could not get above chance performance for this solution.
\end{itemize}

\subsection{Multi-Object Occluded MNIST}
\label{app:mnist}
We used minibatch size 100, and latent dimension $D=10$, yielding two subspaces each of dimension 5.
As the encoder model, we used a three-layer MLP with 50 hidden units per layer and ReLU activations.
We trained for 400 epochs, using Adam~\citep{kingma2014adam} to optimize the encoder, linear classifiers, and discriminators, with separate learning rates for each component chosen via a grid search over $\{ 1e-5, 1e-4, 1e-3 \}$.

\paragraph{Correlated Data Generation.}
We used the default MNIST training and test splits, and held out 10k of the original training examples to form a validation set, yielding 50k, 10k, and 10k examples in the training, validation, and test sets, respectively.
Each digit is first rescaled to be $32 \times 32$ pixels.
The correlated data was generated on-the-fly during training.
Each example in a minibatch was created by: 1) sampling the left-right digit combination (e.g., \{ \texttt{3-3}, \texttt{3-8}, \texttt{8-3}, \texttt{8-8} \}) from a joint distribution encoding the desired correlation; 2) choosing random instances of each of the selected classes (e.g., a random image of a \texttt{3} and a random image of an \texttt{8}); 3) applying occlusions separately to each image; and 4) concatenating the images, yielding a $32 \times 64$ example.
This procedure was performed for each training and test minibatch, yielding a larger amount of data than would be possible with a fixed dataset generated \textit{a priori}.
To generate occlusions, we use the approach from~\citep{chai2021using}, which produces contiguous masks similar to Perlin noise~\citep{perlin2002improving}.
We used gray occlusions to remove a potential ambiguity that exists with black masks (which blend into the MNIST background): a masked \texttt{8} can become identical to a \texttt{3}, so one could not tell whether the image is a noisy \texttt{8} or a clean \texttt{3}.

\begin{figure}[h!]
    \centering
    \begin{subfigure}{.32\textwidth}
    \includegraphics[width=\linewidth]{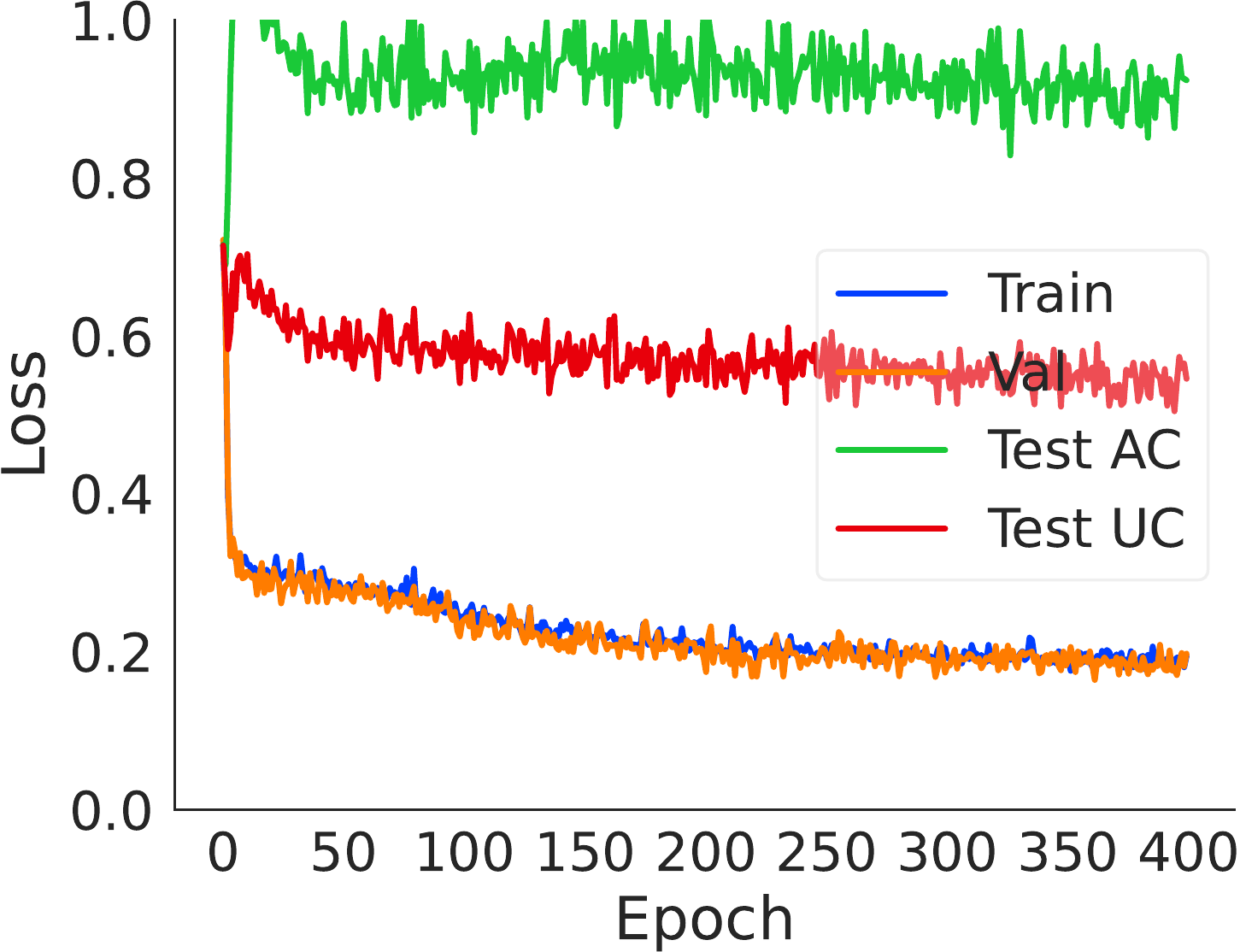}
      \caption{Base}
    \end{subfigure}
    \hfill
    \begin{subfigure}{.32\textwidth}
    \includegraphics[width=\linewidth]{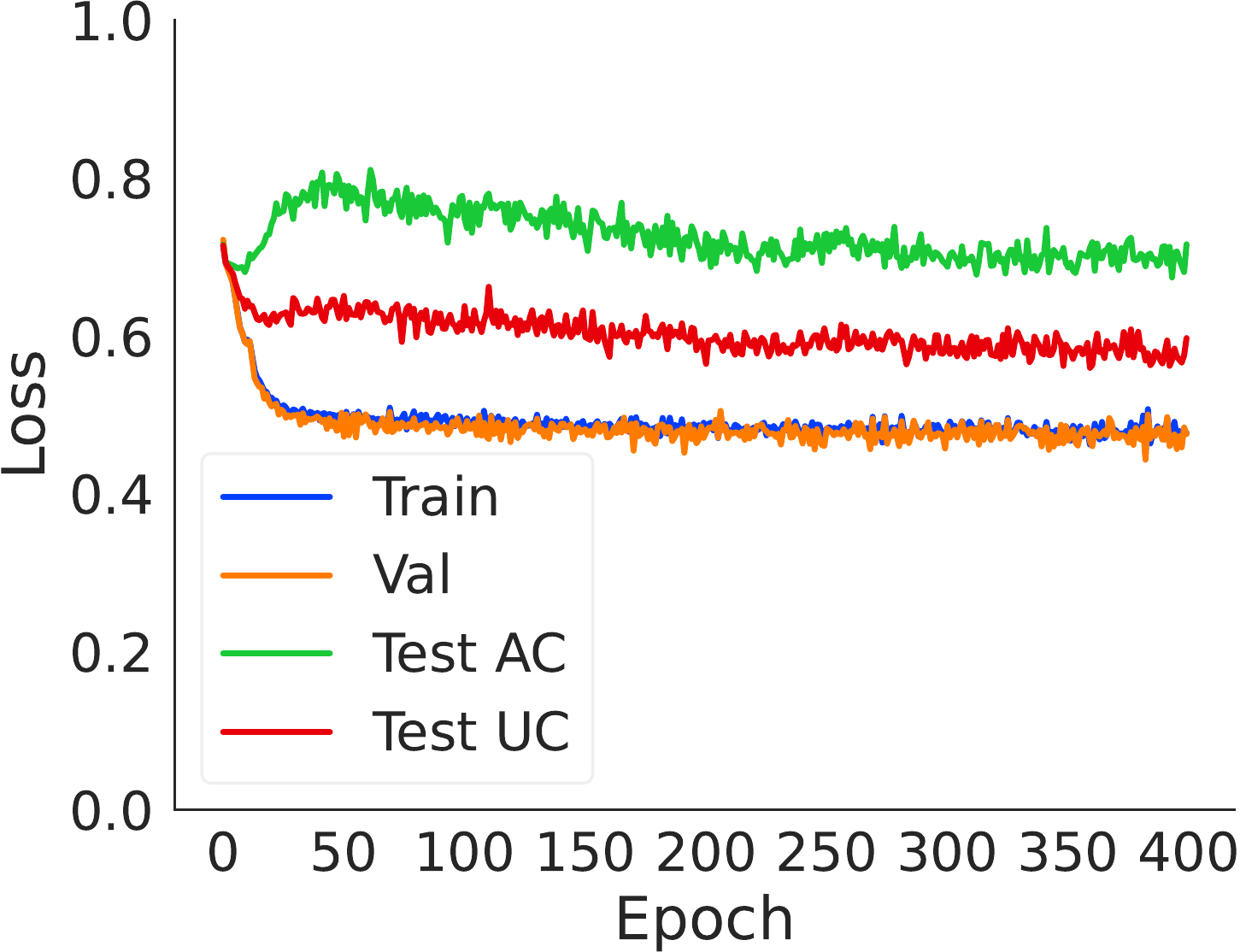}
      \caption{Base + MI}
    \end{subfigure}
    \hfill
    \begin{subfigure}{.32\textwidth}
    \includegraphics[width=\linewidth]{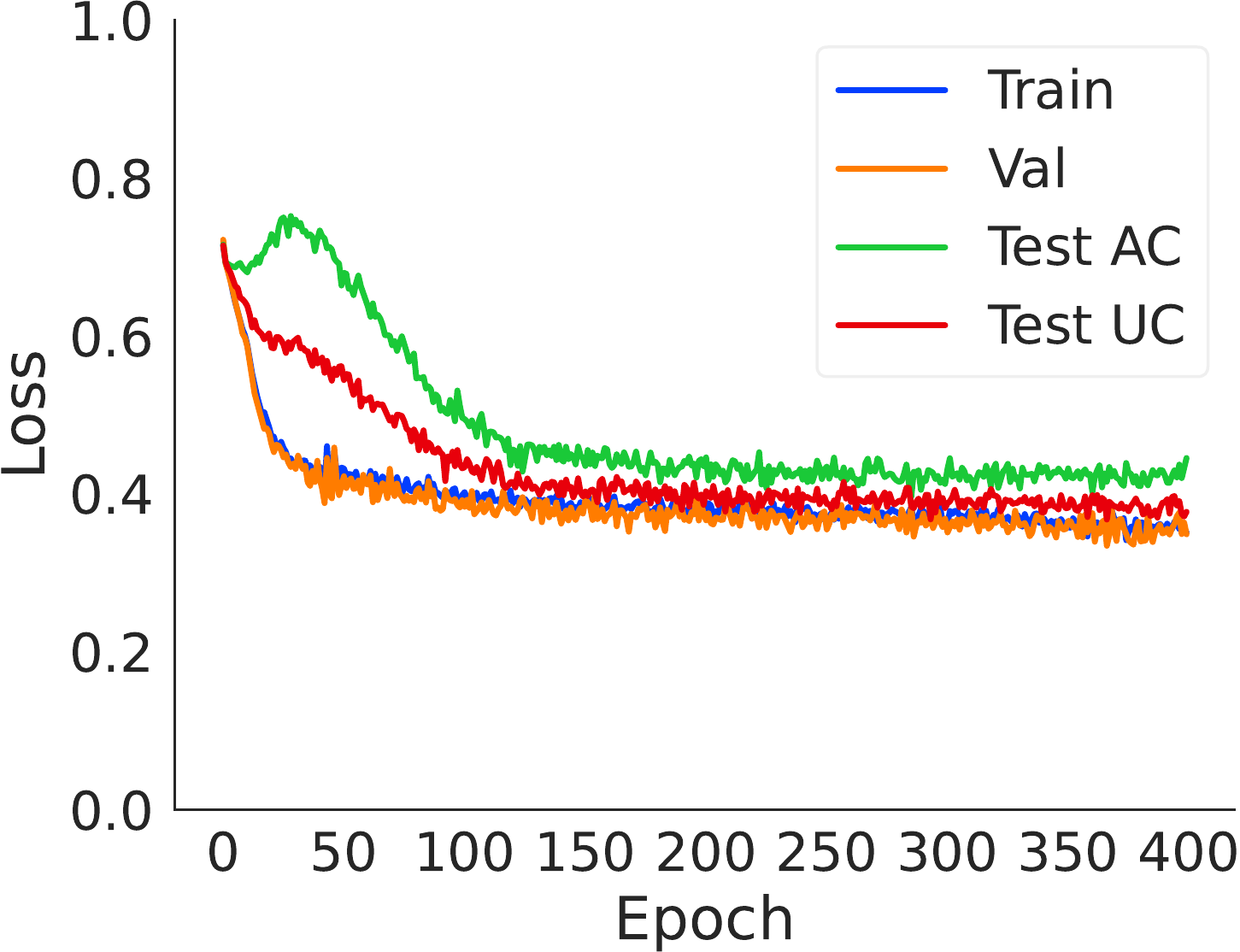}
      \caption{Base + CMI}
    \end{subfigure}
    \caption{Average cross-entropy loss for the left and right digit predictions, under the strongest correlation we consider, $c=0.9$, at noise level 0.6 (where the noise is parameterized by a factor that has range $[0,1]$).
    }
    \label{fig:left-right-f1-losses}
\end{figure}

\begin{figure}[h!]
    \centering
    \begin{subfigure}{.32\textwidth}
    \includegraphics[width=\linewidth]{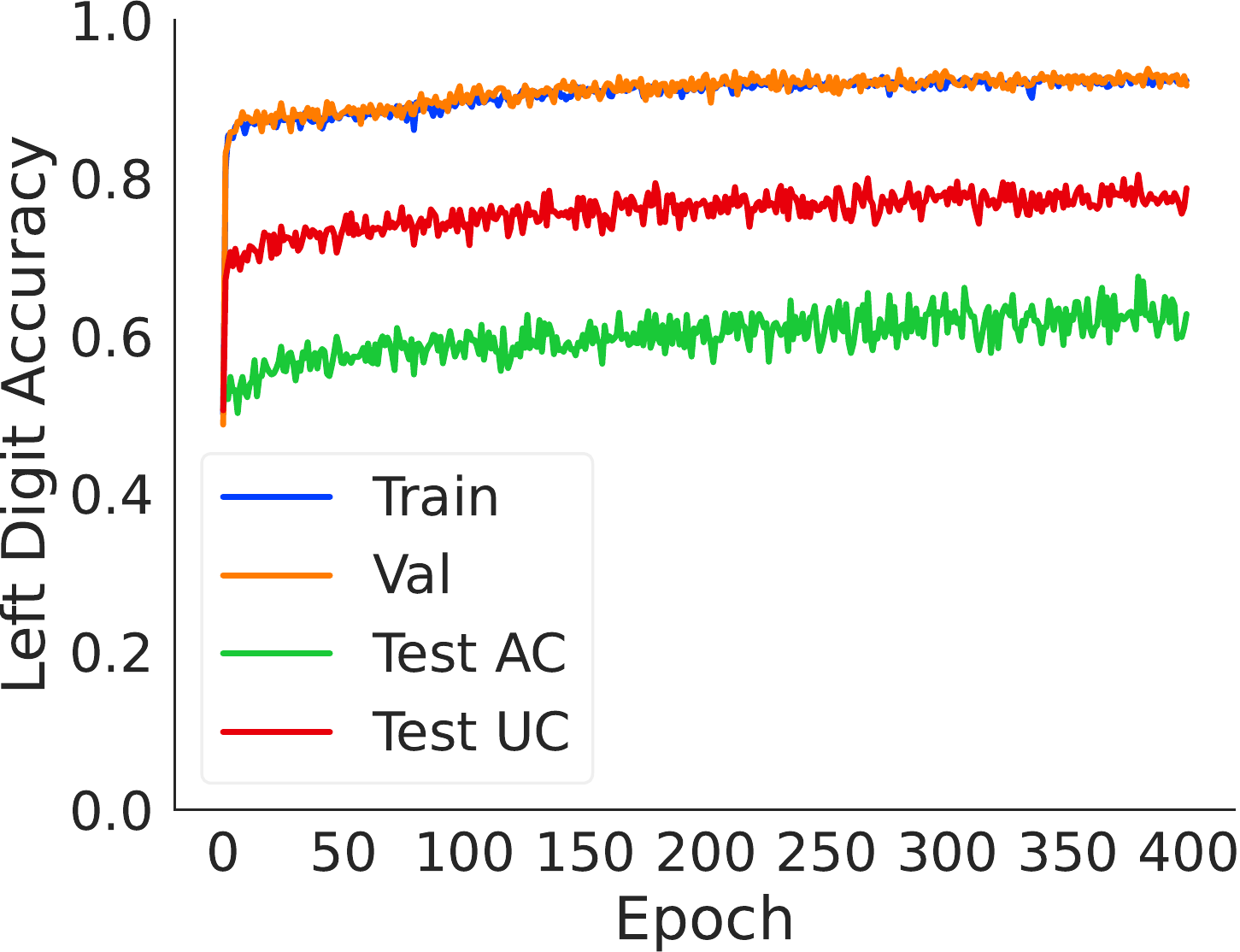}
      \caption{Base}
    \end{subfigure}
    \hfill
    \begin{subfigure}{.32\textwidth}
    \includegraphics[width=\linewidth]{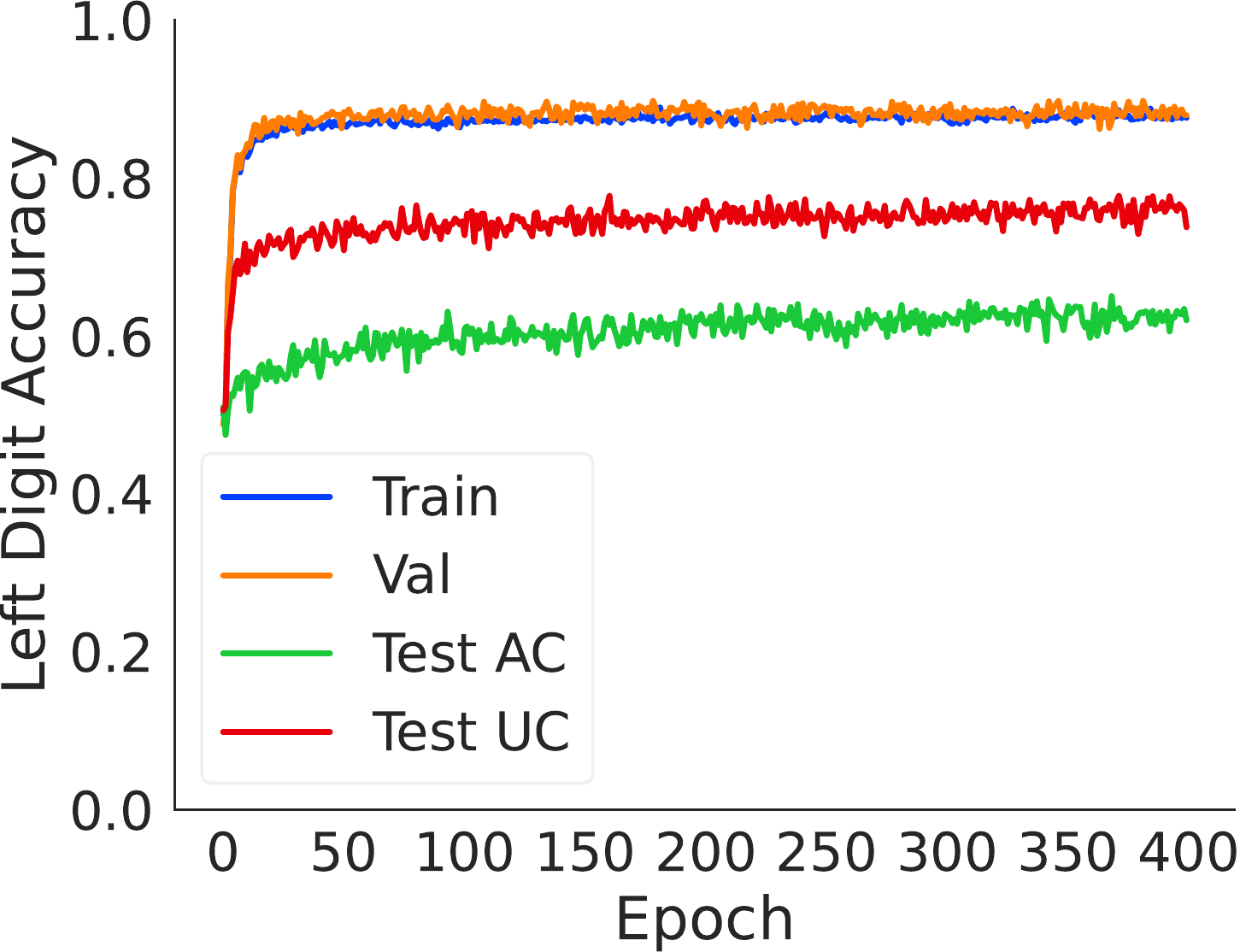}
      \caption{Base + MI}
    \end{subfigure}
    \hfill
    \begin{subfigure}{.32\textwidth}
    \includegraphics[width=\linewidth]{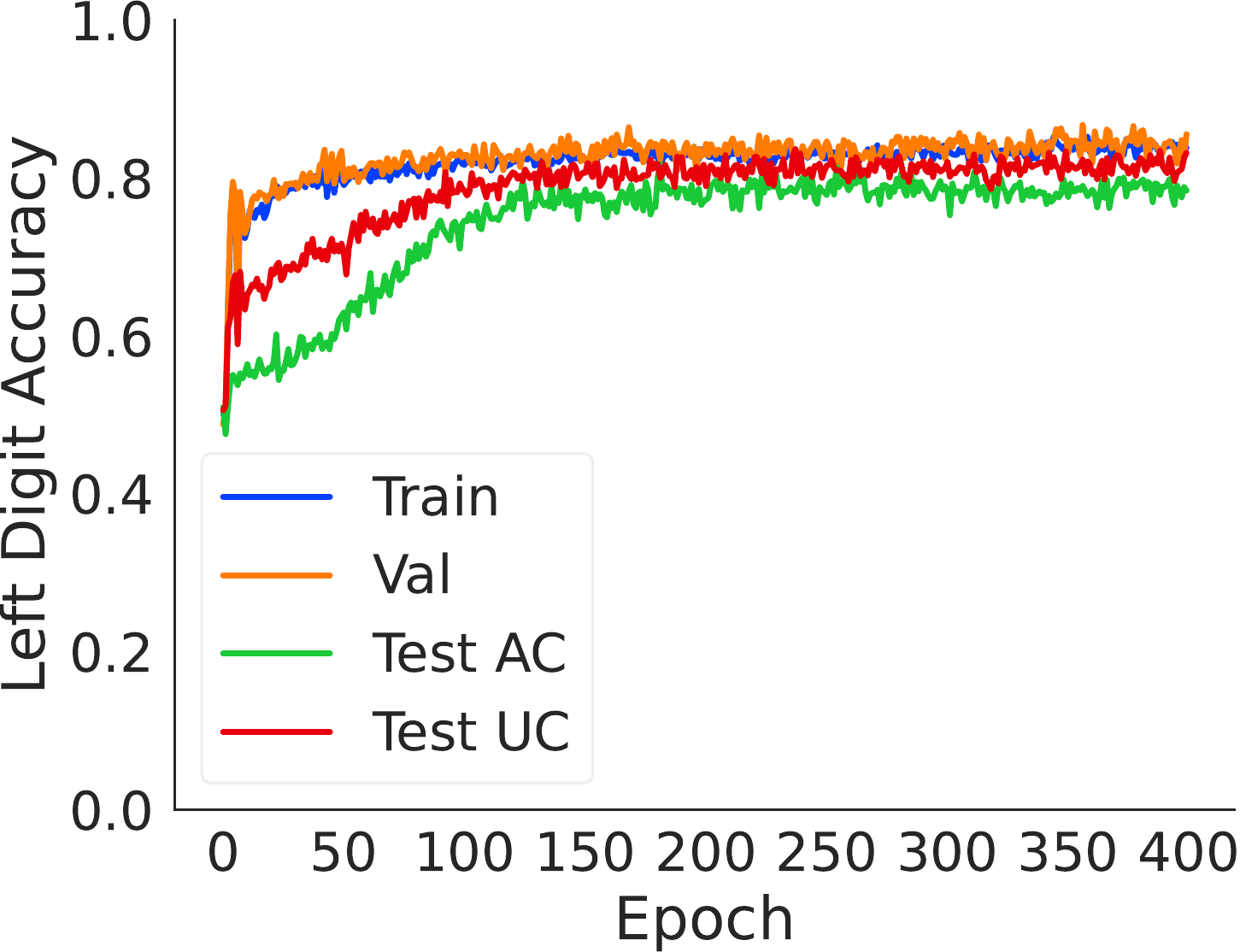}
      \caption{Base + CMI}
    \end{subfigure}
    \caption{Accuracies for the left digit, under the strongest correlation we consider, $c=0.9$, at noise level 0.6 (where the noise is parameterized by a factor that has range $[0,1]$).
    }
    \label{fig:left-right-f1-accs}
\end{figure}

\begin{figure}[h!]
    \centering
    \begin{subfigure}{.32\textwidth}
    \includegraphics[width=\linewidth]{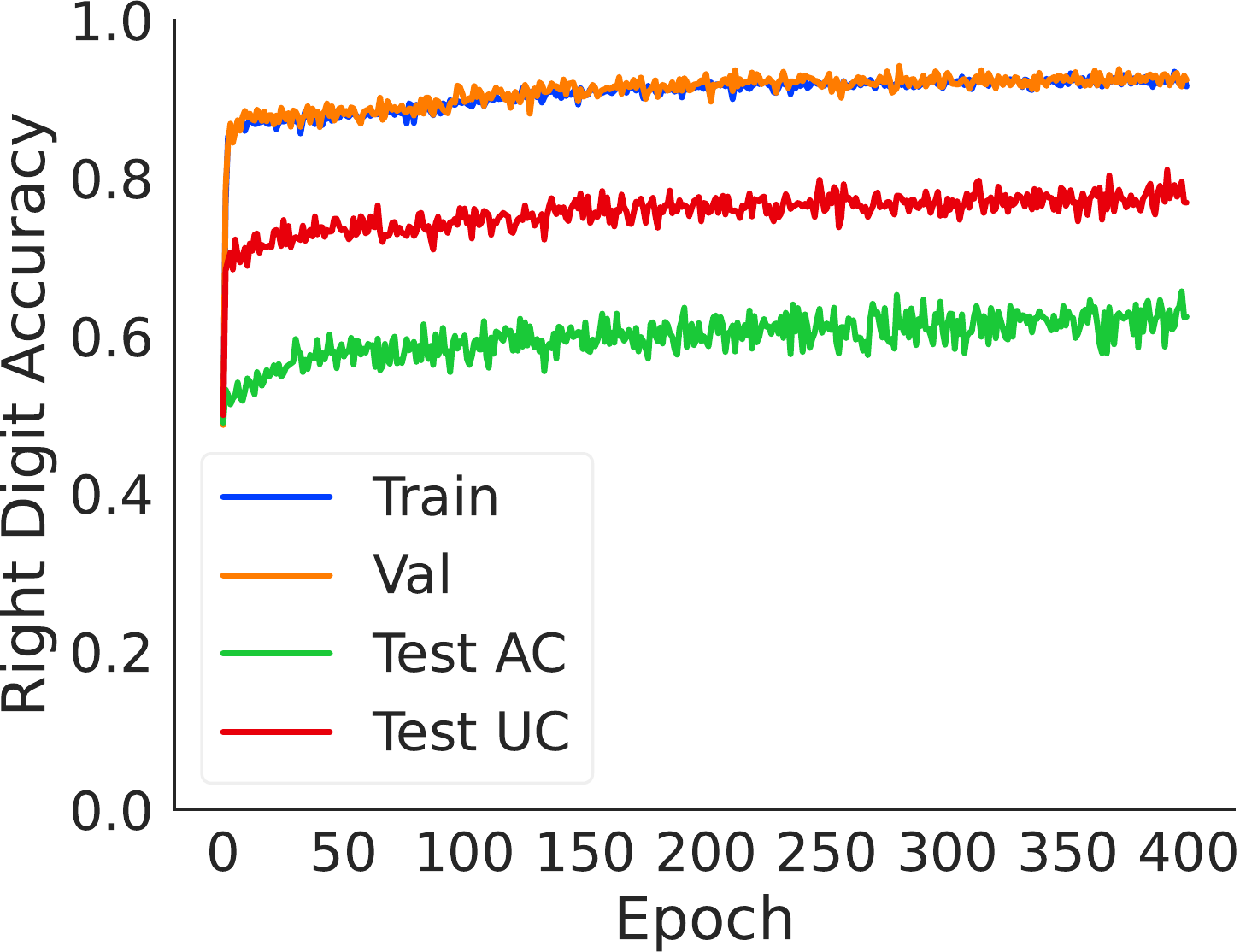}
      \caption{Base}
    \end{subfigure}
    \hfill
    \begin{subfigure}{.32\textwidth}
    \includegraphics[width=\linewidth]{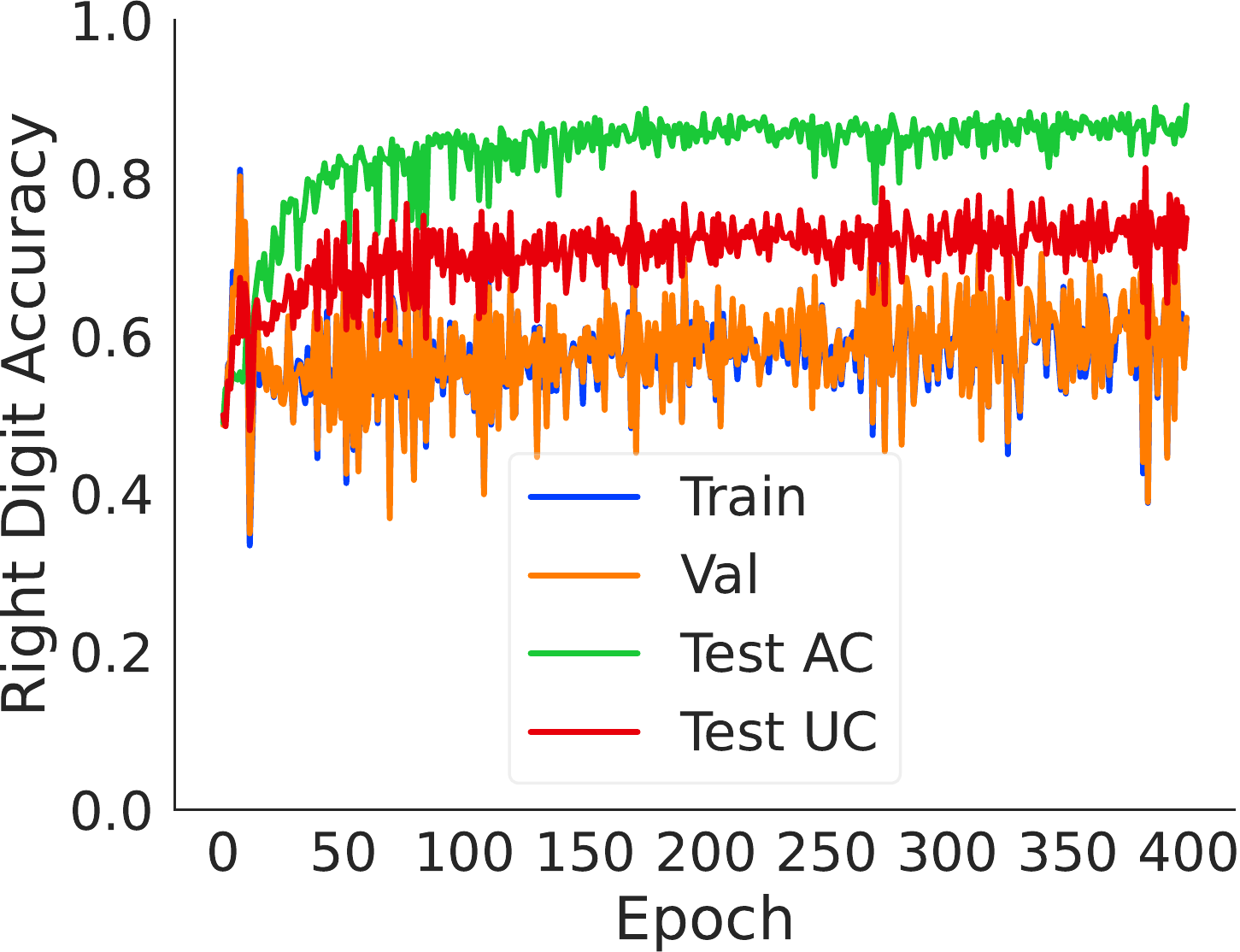}
      \caption{Base + MI}
    \end{subfigure}
    \hfill
    \begin{subfigure}{.32\textwidth}
    \includegraphics[width=\linewidth]{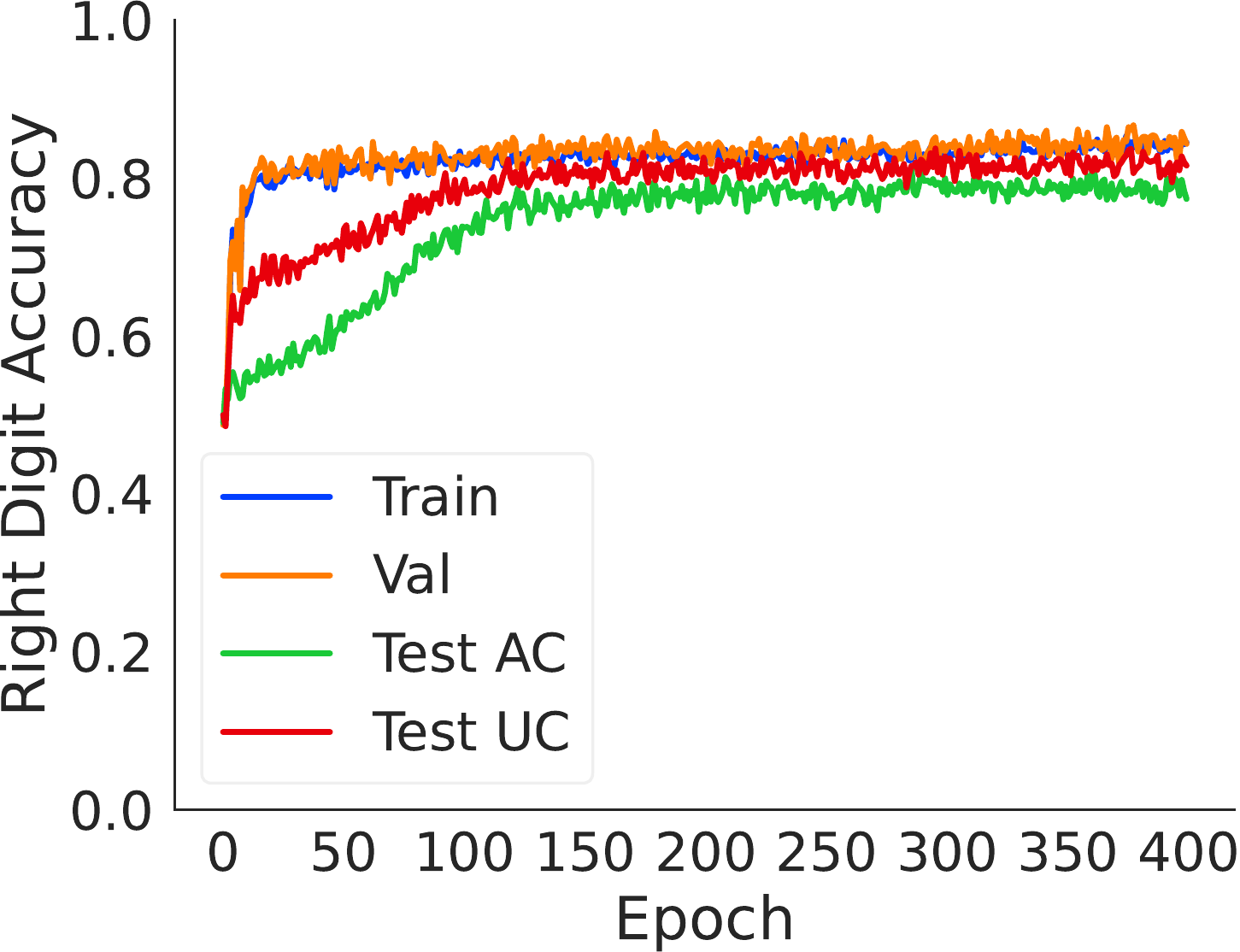}
      \caption{Base + CMI}
    \end{subfigure}
    \caption{Accuracies for the right digit, under the strongest correlation we consider, $c=0.9$, at noise level 0.6 (where the noise is parameterized by a factor that has range $[0,1]$).}
    \label{fig:left-right-f2-accs}
\end{figure}

\subsection{CelebA}
\label{app:celeba}
For all experiments, we used minibatch size 100, and latent dimension $D=10$.
As the encoder model, we used a three-layer MLP with 50 hidden units per layer and ReLU activations.
Similarly to the MNIST setup, we trained for 200 epochs, using Adam to optimize the encoder, linear classifiers, and discriminators.
For each method, we performed a grid search over learning rates $\{ 1e-5, 1e-4, 1e-3 \}$ separately for each of the encoder, discriminator(s), and linear classification heads; we selected the best learning rates based on validation accuracy.

\begin{figure}[h!]
    \centering
    \includegraphics[width=0.23\linewidth]{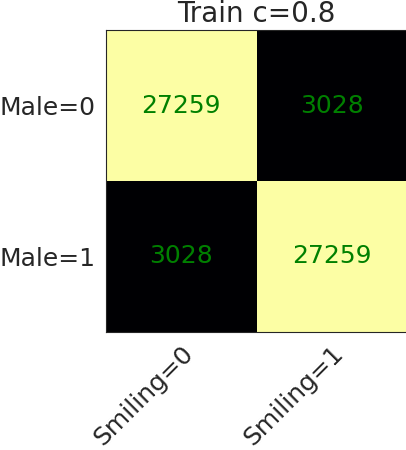}
    \hfill
    \includegraphics[width=0.23\linewidth]{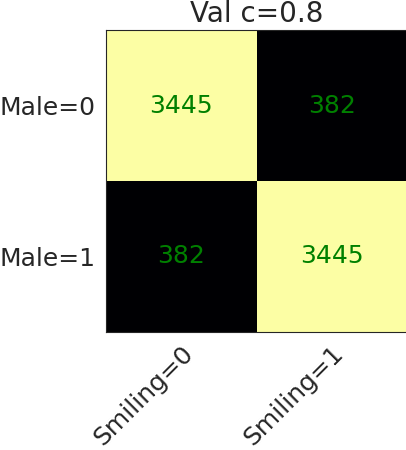}
    \hfill
    \includegraphics[width=0.23\linewidth]{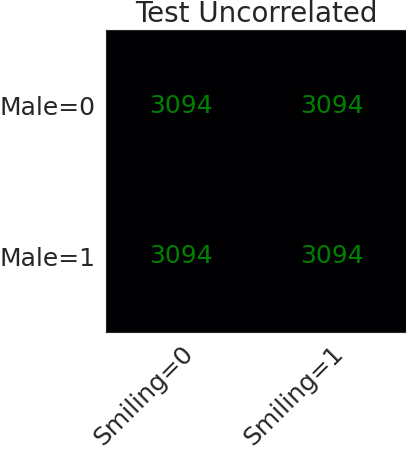}
    \hfill
    \includegraphics[width=0.24\linewidth]{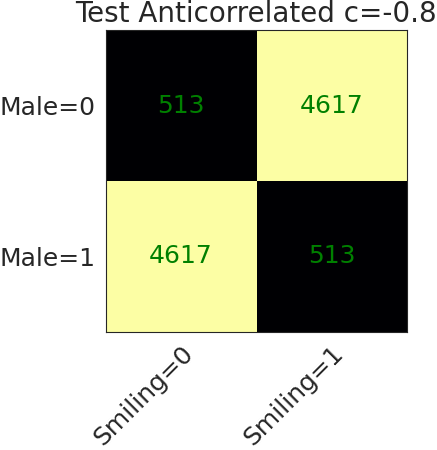}
    \caption{Numbers of examples in the subsampled CelebA datasets for the strongest correlation we consider, $c=0.8.$}
    \label{fig:num-celeba-c8}
\end{figure}

\begin{figure}[h!]
    \centering
    \begin{subfigure}{.32\textwidth}
    \includegraphics[width=\linewidth]{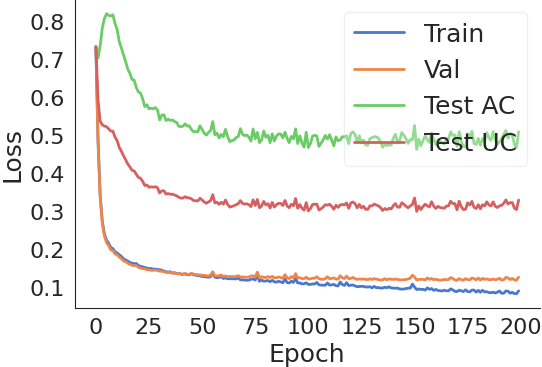}
      \caption{Base}
    \end{subfigure}
    \hfill
    \begin{subfigure}{.32\textwidth}
    \includegraphics[width=\linewidth]{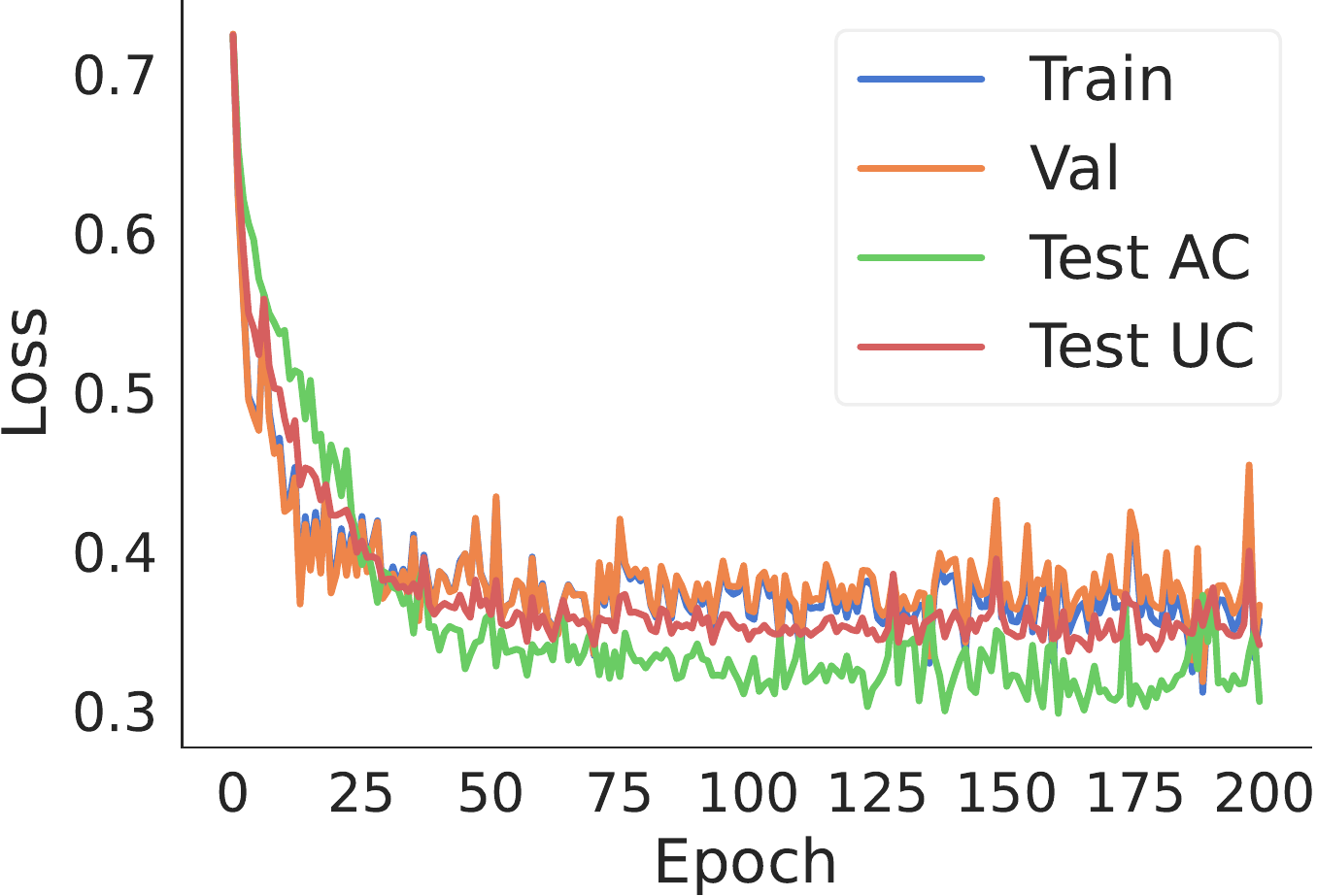}
      \caption{Base + MI}
    \end{subfigure}
    \hfill
    \begin{subfigure}{.32\textwidth}
    \includegraphics[width=\linewidth]{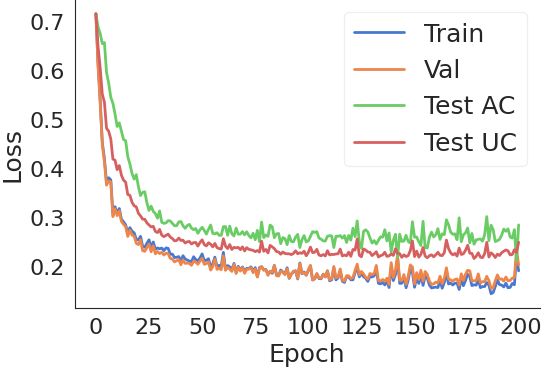}
      \caption{Base + CMI}
    \end{subfigure}
    \caption{Loss curves for each approach on the \texttt{Male-Smiling} CelebA task, under the strongest correlation we consider, $c=0.8$.
    }
    \label{fig:celeba-c8-loss-curves}
\end{figure}

\begin{figure}[h!]
    \centering
    \begin{subfigure}{.32\textwidth}
    \includegraphics[width=\linewidth]{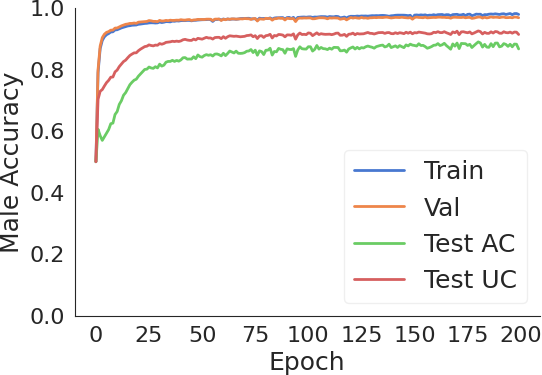}
      \caption{Base}
    \end{subfigure}
    \hfill
    \begin{subfigure}{.32\textwidth}
    \includegraphics[width=\linewidth]{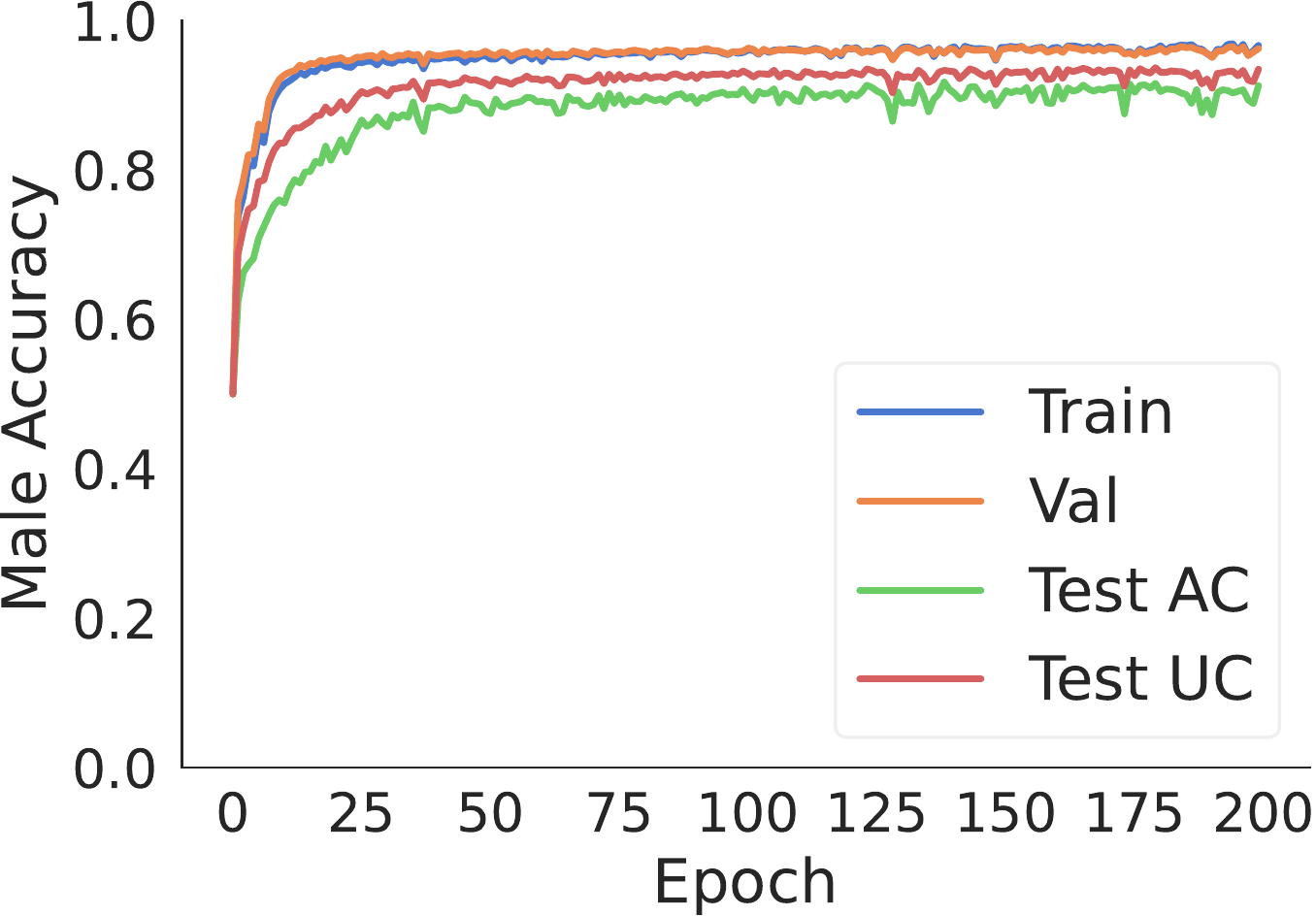}
      \caption{Base + MI}
    \end{subfigure}
    \hfill
    \begin{subfigure}{.32\textwidth}
    \includegraphics[width=\linewidth]{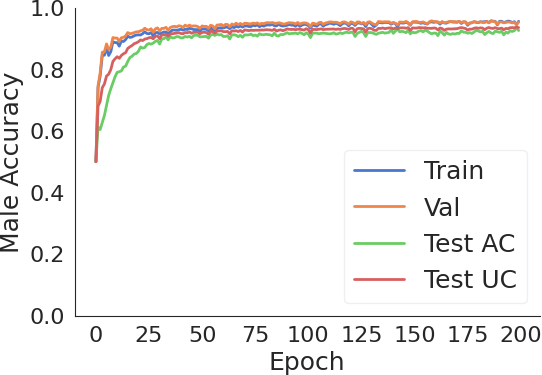}
      \caption{Base + CMI}
    \end{subfigure}
    \caption{Accuracies on the attribute \texttt{Male} for each approach on the \texttt{Male-Smiling} CelebA task, under the strongest correlation we consider, $c=0.8$.}
    \label{fig:celeba-c8-f1-curves}
\end{figure}

\begin{figure}[h!]
    \centering
    \begin{subfigure}{.32\textwidth}
    \includegraphics[width=\linewidth]{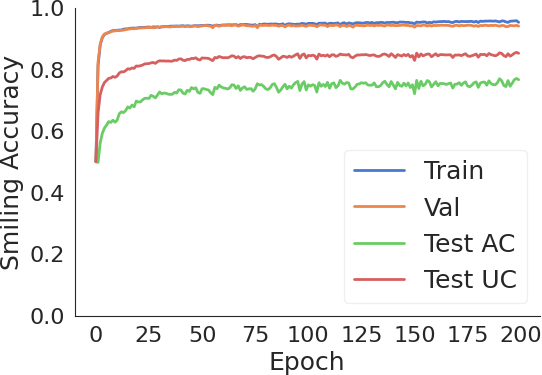}
      \caption{Base}
    \end{subfigure}
    \hfill
    \begin{subfigure}{.32\textwidth}
    \includegraphics[width=\linewidth]{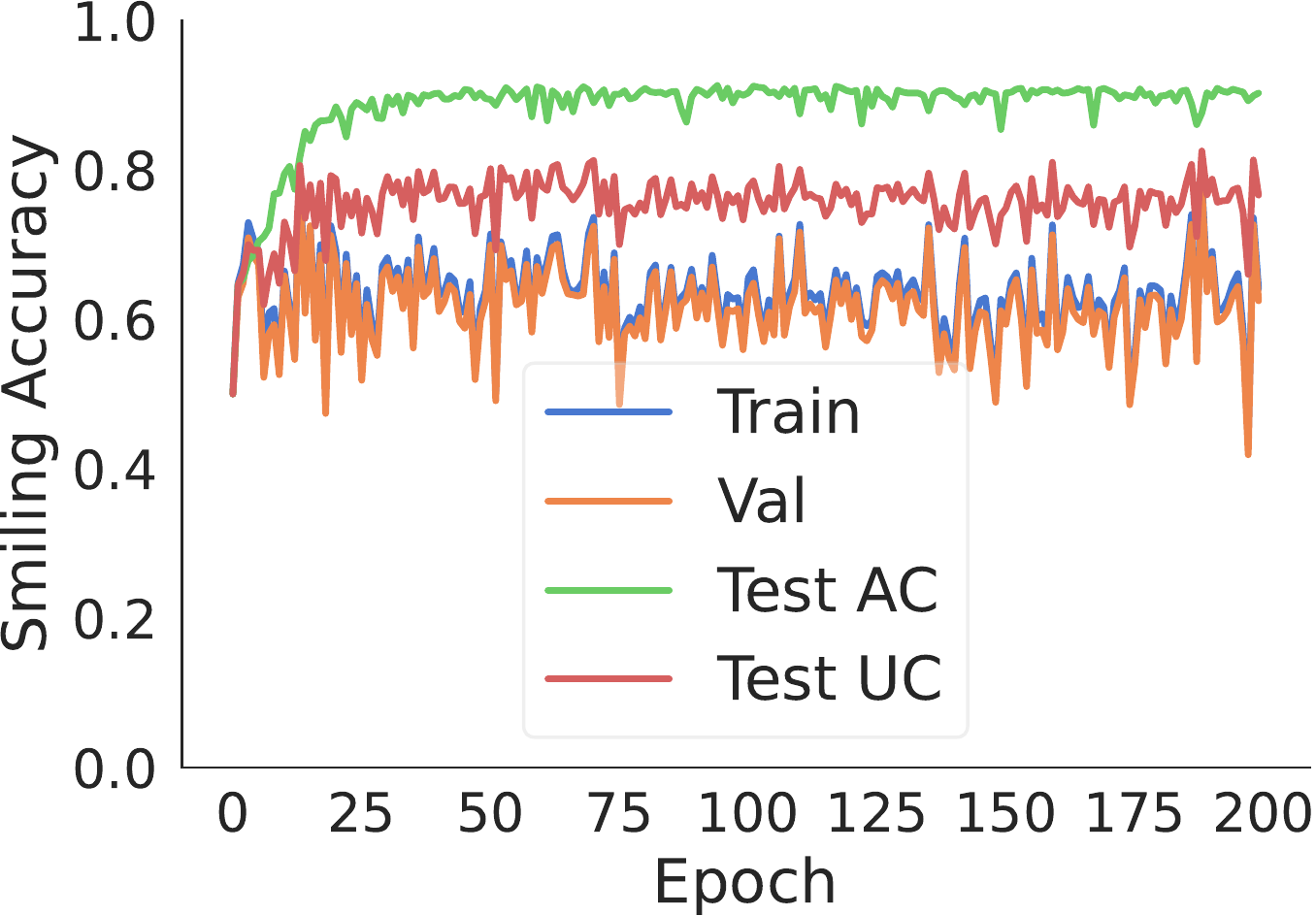}
      \caption{Base + MI}
    \end{subfigure}
    \hfill
    \begin{subfigure}{.32\textwidth}
    \includegraphics[width=\linewidth]{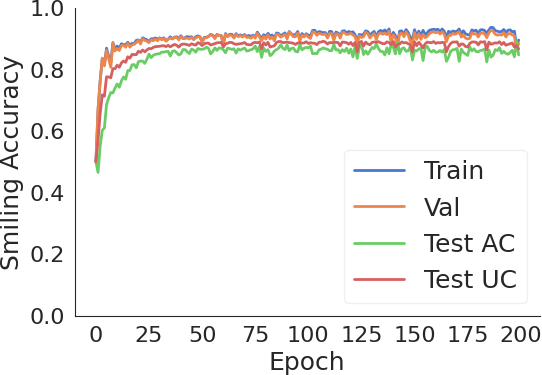}
      \caption{Baes + CMI}
    \end{subfigure}
    \caption{Accuracies on the attribute \texttt{Smiling} for each approach on the \texttt{Male-Smiling} CelebA task, under the strongest correlation we consider, $c=0.8$.
    }
    \label{fig:celeba-c8-f2-curves}
\end{figure}

\begin{figure}[h!]
    \centering
    \begin{subfigure}{.32\textwidth}
    \includegraphics[width=\linewidth]{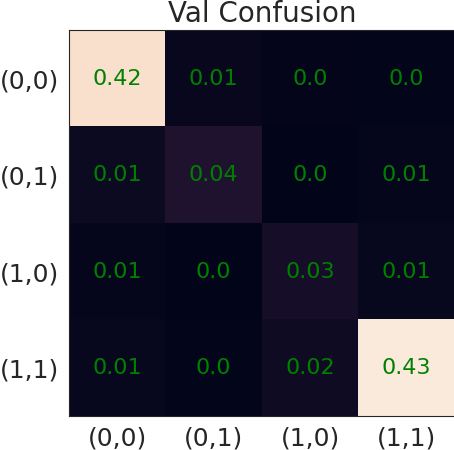}
      \caption{Base}
    \end{subfigure}
    \hfill
    \begin{subfigure}{.32\textwidth}
    \includegraphics[width=\linewidth]{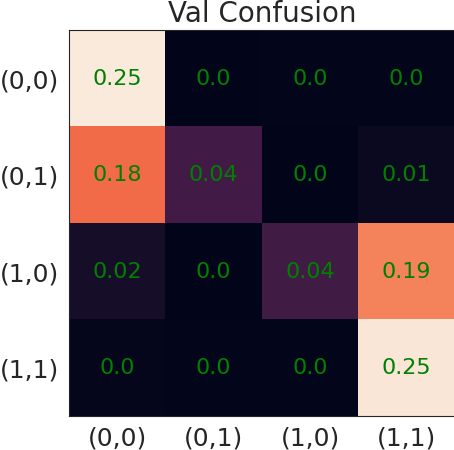}
      \caption{Base + MI}
    \end{subfigure}
    \hfill
    \begin{subfigure}{.32\textwidth}
    \includegraphics[width=\linewidth]{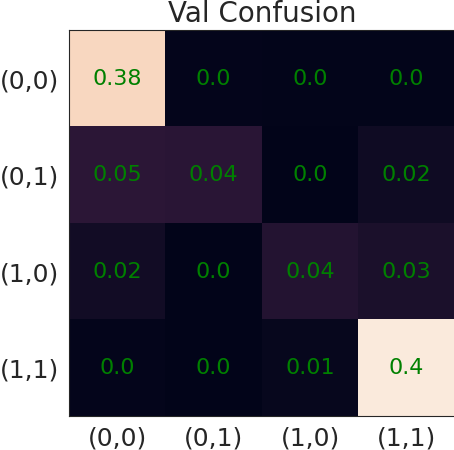}
      \caption{Base + CMI}
    \end{subfigure}
    \caption{Confusion matrices for each approach on the correlated validation set of the \texttt{Male-Smiling} CelebA task, under the strongest correlation we consider, $c=0.8$.}
    \label{fig:celeba-c8-confusion-val}
\end{figure}

\begin{figure}[h!]
    \centering
    \begin{subfigure}{.32\textwidth}
    \includegraphics[width=\linewidth]{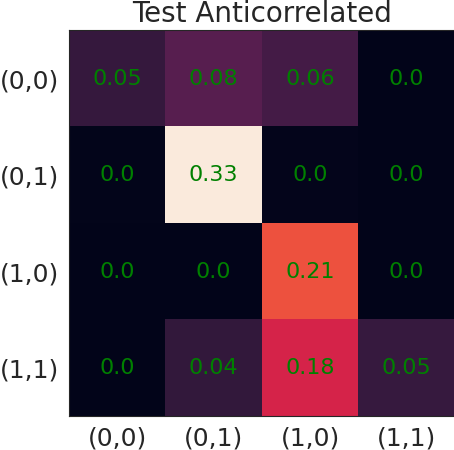}
      \caption{Base}
    \end{subfigure}
    \hfill
    \begin{subfigure}{.32\textwidth}
    \includegraphics[width=\linewidth]{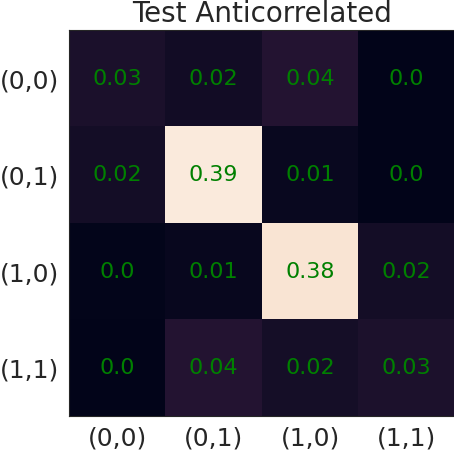}
      \caption{Base + MI}
    \end{subfigure}
    \hfill
    \begin{subfigure}{.32\textwidth}
    \includegraphics[width=\linewidth]{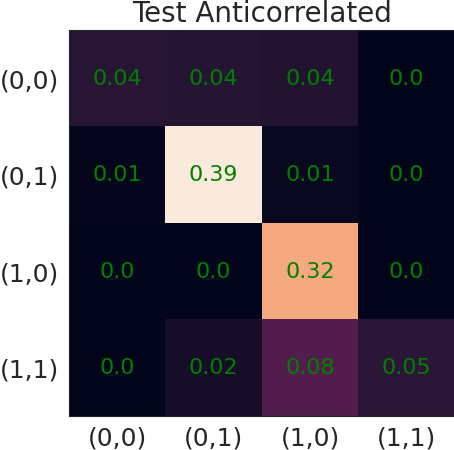}
      \caption{Base + CMI}
    \end{subfigure}
    \caption{Confusion matrices for each approach on the anti-correlated test set of the \texttt{Male-Smiling} CelebA task, under the strongest correlation we consider, $c=0.8$.}
    \label{fig:celeba-c8-confusion-ac}
\end{figure}

\begin{figure}[h!]
    \centering
    \begin{subfigure}{.32\textwidth}
    \includegraphics[width=\linewidth]{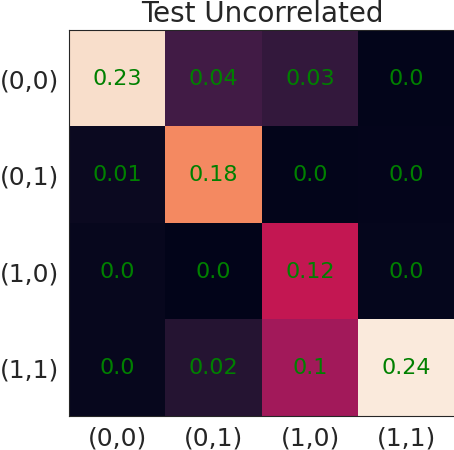}
      \caption{Base}
    \end{subfigure}
    \hfill
    \begin{subfigure}{.32\textwidth}
    \includegraphics[width=\linewidth]{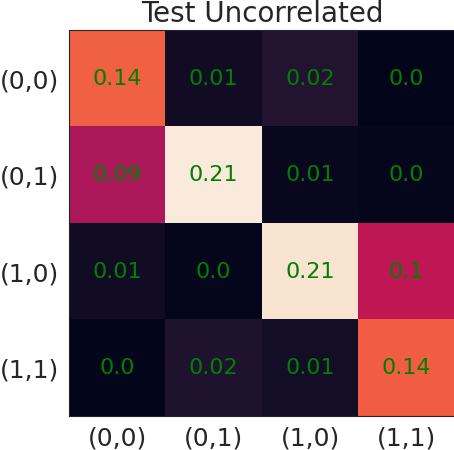}
      \caption{Base + MI}
    \end{subfigure}
    \hfill
    \begin{subfigure}{.32\textwidth}
    \includegraphics[width=\linewidth]{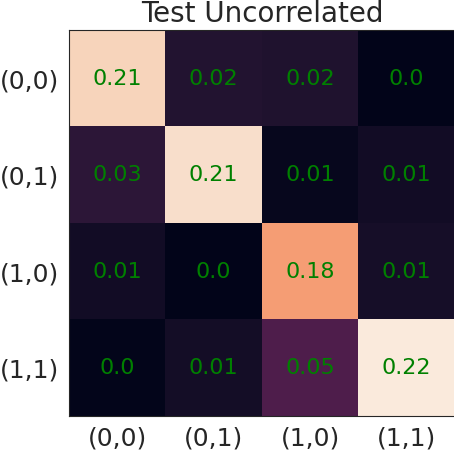}
      \caption{Base + CMI}
    \end{subfigure}
    \caption{Confusion matrices for each approach on the uncorrelated test set of the \texttt{Male-Smiling} CelebA task, under the strongest correlation we consider, $c=0.8$.}
    \label{fig:celeba-c8-confusion-uc}
\end{figure}

\paragraph{Correlated Data Generation.}
We first pre-processed all images by taking a $128 \times 128$ center crop, and then resizing to $64 \times 64$.
Pixel values were normalized to the range $[0,1]$.
We used the original training, validation, and test splits provided with the CelebA dataset.
In order to enforce arbitrary correlations between specific attributes, we subsampled the data such that we retained the maximum possible number of examples in each of the Train/Validation/Anticorrelated Test/Uncorrelated Test splits, while satisfying precisely the desired correlation.
The validation set has the same correlation as the training set, and 
Figure~\ref{fig:num-celeba-c8} shows the number of examples in each of these sets for the strongest correlation we consider, $c=0.8$.
Figures~\ref{fig:celeba-c8-loss-curves}, \ref{fig:celeba-c8-f1-curves}, and \ref{fig:celeba-c8-f2-curves} show the cross-entropy loss and accuracies on each of the factors \texttt{Male} and \texttt{Smiling} (with training correlation 0.8) over the course of optimization, for each of the methods we compare (classification-only, unconditional disentanglement, and conditional disentanglement).
We see that the conditional model substantially outperforms the baselines, with a much smaller gap between validation accuracy and both anti-correlated (AC) and uncorrelated (UC) test accuracies.
Figures~\ref{fig:celeba-c8-confusion-val},~\ref{fig:celeba-c8-confusion-ac} and~\ref{fig:celeba-c8-confusion-uc} show confusion matrices for each method on the correlated validation set, anticorrelated test set, and uncorrelated test set, respectively.
Finally, Tables~\ref{table:subgroup_performance_natural} and \ref{table:subgroup_performance_val} show the prediction error of the models trained with the different objectives for both the combinations that were common and rare during training. These results shows that some attribute combinations (such as the rare non-smiling male faces) are reliably treated incorrectly.

\begin{table}
\centering
\begin{tabular}{l  c  c  c  c}
\toprule
& \multicolumn{2}{c}{\textbf{Common Combinations}} & \multicolumn{2}{c}{\textbf{Rare Combinations}} \\
\midrule
& Female & Male & Female & Male \\
& + & + & + & +  \\
& Non-Smiling & Smiling & Smiling & Non-Smiling \\
\midrule
\textbf{Base} & \textbf{4\%} & \textbf{4\%} & 29\% & 51\% \\ 
\textbf{Base + MI} & 23\% & 28\% & \textbf{12\%} & 31\% \\ 
\textbf{Base + CMI} & 10\% & 9\% & 20\% & \textbf{29\%} \\ 
\bottomrule
\end{tabular}
\vspace{-0.1cm}
\caption{Percentage of incorrect predictions per subgroup for CelebA, evaluated on natural data (e.g., data with naturally-occurring correlations, that has not been subsampled to induce a specific correlation strength), using models trained on correlated data with $c = 0.8$.}
\label{table:subgroup_performance_natural}
\vspace{-0.4cm}
\end{table}

\begin{table}%[h]
\centering
\begin{tabular}{l  c  c  c  c}
\toprule
& \multicolumn{2}{c}{\textbf{Common Combinations}} & \multicolumn{2}{c}{\textbf{Rare Combinations}} \\
\midrule
& Female & Male & Female & Male \\
& + & + & + & +  \\
& Non-Smiling & Smiling & Smiling & Non-Smiling \\
\midrule
\textbf{Base} & \textbf{4\%} & \textbf{5\%} & 33\% & 49\%\\ 
\textbf{Base + MI} & 24\% & 28\% & \textbf{11}\% & 26\%\\ 
\textbf{Base + CMI} & 9\% & 9\% & 19\% & \textbf{25\%}\\ 
\bottomrule
\end{tabular}
\vspace{-0.1cm}
\caption{Percentage of incorrect predictions per subgroup for CelebA, evaluated on validation data ($c = 0.8$), using models trained on correlated data with $c = 0.8$.}
\label{table:subgroup_performance_val}
\vspace{-0.4cm}
\end{table}

\clearpage
\subsection{Disentanglement Metrics}
\label{app:dis_metrics}
We evaluated common disentanglement metrics~\citep{locatello2019challenging} on uncorrelated test data using models trained on correlated data.
We performed this analysis for two of our datasets and found in both cases that \oCMI reached better scores compared to the other objectives for almost all metrics. 

\textbf{Toy Classification}: Disentanglement results for the toy classification task with ten attributes are shown in Table~\ref{table:dis_metrics_toy}.
We obtained similar results for two and four attributes, which are not reported for brevity.

\textbf{CelebA}: Since the disentanglement metrics require that the factors of variation are each encoded in one-dimensional subspaces, we set latent dimension $D = 2$ for this experiment.
In Table~\ref{table:dis_metrics_celeba}, we report the average and 68\% confidence intervals for five models trained on data with correlation level 0.8.

\begin{table}[htbp]
\centering
\begin{tabular}{l  c  c  c}
\toprule
\textbf{Metric} & \textbf{Base} & \textbf{Base+MI} & \textbf{Base+CMI} \\
\midrule
IRS~\citep{suter2019robustly}  $\uparrow$ & 0.377& 0.573& \textbf{0.605}\\
SAP~\citep{kumar2017variational} $\uparrow$ & 0.118& 0.470& \textbf{0.477}\\
MIG~\citep{chen2018isolating} $\uparrow$ & 0.179& 0.939& \textbf{0.975}\\
DCI Disentanglement~\citep{eastwood2018framework} $\uparrow$ & 0.413 & 0.980& \textbf{0.998}\\
Beta-VAE~\citep{higgins2016beta} $\uparrow$ & 0.996 & 1 & 1\\
Factor-VAE~\citep{kim2018disentangling} $\uparrow$ & 1 & 1& 1\\
Gaussian Total Correlation $\downarrow$ & 10.073 & 0.485& \textbf{0.025} \\
Gaussian Wasserstein Corr $\downarrow$ & 12.905& 0.373& \textbf{0.027} \\
Gaussian Wasserstein Corr Norm $\downarrow$ & 0.866& 0.037 & \textbf{0.002}\\
Mutual Info Score $\downarrow$ & 0.975 & 0.197 & \textbf{0.149} \\
\bottomrule
\end{tabular}
\vspace{-0.1cm}
\caption{\textbf{Disentanglement metrics for toy classification with ten attributes.} Metrics are evaluated on the uncorrelated test set. Bold font indicates model with best disentanglement score.}
\label{table:dis_metrics_toy}
\vspace{-0.4cm}
\end{table}

\begin{table}[htbp]
\centering
\begin{tabular}{l  c  c  c}
\toprule
\textbf{Metric} & \textbf{Base} & \textbf{Base+MI} & \textbf{Base+CMI} \\
\hline
IRS $\uparrow$ & 0.524 $\pm$ 0.043 & \textbf{0.548 $\pm$ 0.038} &  0.531 $\pm$ 0.041 \\
SAP $\uparrow$ & 0.306 $\pm$ 0.003 & 0.296 $\pm$ 0.046 & \textbf{ 0.389 $\pm$ 0.005 }\\
MIG $\uparrow$ & 0.506 $\pm$ 0.01 & 0.455 $\pm$ 0.074 & \textbf{ 0.674 $\pm$ 0.007 }\\
DCI Disentanglement $\uparrow$ & 0.46 $\pm$ 0.009 & 0.596 $\pm$ 0.038 & \textbf{ 0.807 $\pm$ 0.023 }\\
Beta-VAE $\uparrow$ & 1.0 $\pm$ 0.0 & 1.0 $\pm$ 0.0 & \textbf{ 1.0 $\pm$ 0.0 }\\
Factor-VAE $\uparrow$ & 1.0 $\pm$ 0.0 & 0.999 $\pm$ 0.003 & \textbf{ 1.0 $\pm$ 0.0 }\\
Gaussian Total Correlation $\downarrow$ & 0.222 $\pm$ 0.012 & 0.056 $\pm$ 0.061 & \textbf{ 0.011 $\pm$ 0.003 }\\
Gaussian Wasserstein Corr $\downarrow$ & 0.351 $\pm$ 0.039 & 0.01 $\pm$ 0.009 & \textbf{ 0.002 $\pm$ 0.001 }\\
Gaussian Wasserstein Corr Norm $\downarrow$ & 0.098 $\pm$ 0.005 & 0.006 $\pm$ 0.004 & \textbf{ 0.005 $\pm$ 0.001 }\\
Mutual Info Score $\downarrow$ & 0.302 $\pm$ 0.022 & 0.111 $\pm$ 0.052 & \textbf{ 0.042 $\pm$ 0.006 }\\
\bottomrule
\end{tabular}
\caption{\textbf{Disentanglement metrics for CelebA.} Metrics are evaluated on the uncorrelated test set. Bold font indicates model with best disentanglement score.} % average of 5 models
\label{table:dis_metrics_celeba}
\vspace{-0.4cm}
\end{table}

\subsection{Weakly Supervised Setting}
\label{app:weakly}
For the fully supervised CelebA experiment, labels for both attributes were available for all 10260 images. For the weakly supervised setting, we reduced the number of labels to 5130 (50\% of the labels of the fully supervised dataset), 2565 (25\%), 1026 (10\%), or 513 (5\%) for each attribute. %, or 205 (2\%).
This implies that some images had both labels, some had only one label and some images had no labels (for example when using 50\% of the labels the distinction is as follows: 25\% of the images had both labels; 25\% had only labels for attribute 1; 25\% had only labels for attribute 2; and 25\% had no labels). 
The three objectives can be applied to these weakly supervised settings. For \oBase, the cross-entropy loss for each attribute was computed only for the images that had labels for the corresponding attribute. For \oMI no labels are required for the unconditional shuffling; thus this objective can be applied even for the images without labels. For \oCMI, our method shuffles only images that have the same value for a given attribute. This also works if the labels of the other attribute are missing.
We used the same training parameters as for the supervised experiment, except for increasing the number of training epochs (up to 1200 epochs) and adapting the minibatch size to the number of labels. In Figure~\ref{fig:celeba-weakly} we report the average and 68\% confidence intervals over three runs with different seeds.

\section{Algorithms}
\label{app:algorithms}

In this section, we provide formal descriptions of the baseline approaches we use.
Algorithm~\ref{alg:cls-only} describes the classification-only baseline, that trains separate linear classifiers to predict attributes $\bolds_k$ from the corresponding latent subspaces $\boldz_k$.
Algorithm~\ref{alg:disentanglement-uncond-mi} and Algorithm~\ref{alg:disentanglement-uncond-mi-train-disc} describe the unconditional disentanglement baseline, that adversarially minimizes the discrepancy between samples from the joint distribution $p(\boldz_1, \dots, \boldz_k)$ and the product of marginals $p(\boldz_1) \cdots p(\boldz_k)$.
Algorithm~\ref{alg:disentanglement-cond-mi-train-disc} describes the discriminator training loop for the CMI minimization approach from Section~\ref{sec:method}.

\begin{algorithm}[htbp]
	\caption{Supervised Learning on Subspaces
	}
	\label{alg:cls-only}
	\begin{algorithmic}[1]
	    \State \textbf{Input:} $\{ \boldphi_1, \dots, \boldphi_K \}$, initial parameters for $K$ linear classifiers $R_1, \dots, R_K$
        \State \textbf{Input:} $\boldtheta$, initial parameters for the encoder $f$
        \State \textbf{Input:} $\alpha, \beta$ learning rates for training the encoder and linear classifiers
	    \While {true}
	        \State $(\boldx, \{\bolds_k\}_{k=1}^K) \sim \mathcal{D}_{\text{Train}}$  \algorithmiccomment{Sample a minibatch of data with attribute labels}
	        \State $\boldz \gets f_{\boldtheta}(\boldx)$  \algorithmiccomment{Forward pass through the encoder}
        	\State $\{ \boldz_k \}_{k=1}^K \gets \text{SplitSubspaces}(\boldz, k)$  \algorithmiccomment{Partition the latent space into $k$ subspaces}
	        \State $L \gets \sum_{k=1}^K L_{\text{cls}}(R_k(\boldz_k ; \boldphi_k), \bolds_k)$ \algorithmiccomment{Cross-entropy for each attribute}
	       \State $\boldtheta \gets \boldtheta - \alpha \nabla_{\boldtheta} L$  \algorithmiccomment{Update encoder parameters}
	       \State $\boldphi_k \gets \boldphi_k - \beta \nabla_{\boldphi_k} L \quad , \quad \forall k \in \{1, \dots, K \}$  \algorithmiccomment{Update classifier parameters}
	   \EndWhile
	\end{algorithmic}
\end{algorithm}

\begin{algorithm}[htbp]
	\caption{Learning Unconditionally Disentangled Subspaces
	--- Training the Encoder}
	\label{alg:disentanglement-uncond-mi}
	\begin{algorithmic}[1]
	    \State \textbf{Input:} $\{ \boldphi_1, \dots, \boldphi_K \}$, initial parameters for $K$ linear classifiers $R_1, \dots, R_K$
        \State \textbf{Input:} $\boldtheta$, initial parameters for the encoder $f$
        \State \textbf{Input:} $\alpha, \beta$ learning rates for training the encoder and linear classifiers
	    \While {true}
	        \State $(\boldx, \{\bolds_k\}_{k=1}^K) \sim \mathcal{D}_{\text{Train}}$  \algorithmiccomment{Sample a minibatch of data with attribute labels}
	        \State $\boldz \gets f_{\boldtheta}(\boldx)$  \algorithmiccomment{Forward pass through the encoder}
        	\State $\{ \boldz_k \}_{k=1}^K \gets \text{SplitSubspaces}(\boldz, k)$  \algorithmiccomment{Partition the latent space into $k$ subspaces}
	        \State $L \gets \sum_{k=1}^K L_{\text{cls}}(R_k(\boldz_k ; \boldphi_k), \bolds_k)$ \algorithmiccomment{Cross-entropy for each attribute}
            \State $\boldz' \sim p(\boldz_1) p(\boldz_2) \cdots p(\boldz_k)$ \algorithmiccomment{Samples w/ batchwise-shuffled subspaces}
            \State $L \gets L + \log \left( 1 - D_{\boldomega}(\boldz') \right) + \log \left( D_{\boldomega}(\boldz) \right)$ \algorithmiccomment{Add adversarial loss}
	        \State $\boldtheta \gets \boldtheta - \alpha \nabla_{\boldtheta} L$  \algorithmiccomment{Update encoder parameters}
	       \State $\boldphi_k \gets \boldphi_k - \beta \nabla_{\boldphi_k} L \quad , \quad \forall k \in \{1, \dots, K \}$  \algorithmiccomment{Update classifier parameters}
	   \EndWhile
	\end{algorithmic}
\end{algorithm}

\begin{algorithm}[htbp]
	\caption{Learning Unconditionally Disentangled Subspaces
	--- Training the Discriminator}
	\label{alg:disentanglement-uncond-mi-train-disc}
	\begin{algorithmic}[1]
	    \State \textbf{Input:} $\boldomega$, initial parameters for the discriminator $D$
	    \State \textbf{Input:} $\gamma$, learning rate for training the discriminator
	    \While {true}
	        \State $(\boldx, \{\bolds_k\}_{k=1}^K) \sim \mathcal{D}_{\text{Train}}$  \algorithmiccomment{Sample a minibatch of data with attribute labels}
	        \State $\boldz \gets f_{\boldtheta}(\boldx)$  \algorithmiccomment{Forward pass through the encoder}
        	\State $\{ \boldz_k \}_{k=1}^K \gets \text{SplitSubspaces}(\boldz, k)$  \algorithmiccomment{Partition the latent space into $k$ subspaces}
            \State $\boldz' \sim p(\boldz_1) p(\boldz_2) \cdots p(\boldz_k)$ \algorithmiccomment{Samples w/ batchwise-shuffled subspaces}
            \State $L \gets L + \log \left( D_{\boldomega}(\boldz') \right) + \log \left(1 -  D_{\boldomega}(\boldz) \right)$ \algorithmiccomment{Add adversarial loss}
	        \State $\boldomega \gets \boldomega - \gamma \nabla_{\boldomega} L$  \algorithmiccomment{Update discriminator parameters}
	   \EndWhile
	\end{algorithmic}
\end{algorithm}

\begin{algorithm*}[htbp]
	\caption{Learning Conditionally Disentangled Subspaces Adversarially -- Training the Discriminator}
	\label{alg:disentanglement-cond-mi-train-disc}
	\begin{algorithmic}[1]
	    \State \textbf{Input:} $\boldomega$, initial parameters for the discriminator $D$
	    \State \textbf{Input:} $\gamma$, learning rate for training the discriminator
	    \While {true}
	        \State $(\boldx, \{\bolds_k\}_{k=1}^K) \sim \mathcal{D}_{\text{Train}}$  \algorithmiccomment{Sample a minibatch of data with attribute labels}
	        \State $\boldz \gets f_{\boldtheta}(\boldx)$  \algorithmiccomment{Forward pass through the encoder}
        	\State $\{ \boldz_k \}_{k=1}^K \gets \text{SplitSubspaces}(\boldz, k)$  \algorithmiccomment{Partition the latent space into $K$ subspaces}
        	\State $L \gets 0$  \algorithmiccomment{$L$ will accumulate the losses over all subspaces}
	        \For{$k \in \{ 1, \dots, K \}$}
	            \State $\boldz' \sim p(\boldz_1, \dots \boldz_K \mid \bolds_k)$ \algorithmiccomment{Samples from the joint distribution}
	            \State $\boldz'' \sim p(\boldz_k \mid \bolds_k) p(\boldz_{-k} \mid \bolds_k)$ \algorithmiccomment{Samples w/ batchwise-shuffled subspaces}
	           \State $L \gets L + \log \left( D_{\boldomega}(\boldz'') \right) + \log \left( 1 - D_{\boldomega}(\boldz') \right)$ \algorithmiccomment{Add adversarial loss}
	        \EndFor
	       \State $\boldomega \gets \boldomega - \gamma \nabla_{\boldomega} L$  \algorithmiccomment{Update discriminator parameters}
	   \EndWhile
	\end{algorithmic}
\end{algorithm*}

\clearpage

\section{Proof of Proposition~\ref{prop:mi}}
\label{app:proofs}

\textbf{Proposition 3.1}\ \ \textit{If $\text{I}(s_1; s_2) > 0$, then enforcing $\text{I}(z_1; z_2)=0$ means that $\text{I}(z_k; s_k) < H(s_k)$ for at least one $k$.}
\begin{proof}
Assume that $I(s_1; s_2) > 0$ and at the same time $I(z_k; s_k) = H(s_k)$ (i.e., we are proving by contradiction). Since $I(z_1; s_1) = H(s_1)$, we have $H(s_1 \mid z_1) = 0$ and with $H(s_1 \mid z_1) = H(s_1 \mid z_1, s_2) + I(s_1; s_2 \mid z_1)$ (both non-negative), it follows that $H(s_1 \mid z_1, s_2) = I(s_1; s_2 \mid z_1) = 0$. Since for the interaction information, by definition $I(s_1; s_2; z_1) = I(s_1; s_2) - I(s_1; s_2 \mid z_1)$, and $I(s_1; s_2 \mid z_1)=0$, we have $I(s_1; s_2; z_1) = I(s_1; s_2) > 0$. Since we also assume $H(s_2 \mid z_2) = 0$, we also have $I(s_1; s_2; z_2) = I(s_1; s_2) > 0$.

We can use this to compute the fourth order interaction information $I(s_1; s_2; z_1; z_2)$. By definition, we have $I(s_1; s_2; z_1; z_2) = I(s_1; s_2; z_1) - I(s_1; s_2; z_1 \mid z_2)$. We just showed that $I(s_1; s_2; z_1) = I(s_1; s_2)$, and therefore we have $I(s_1; s_2; z_1 \mid z_2) = I(s_1; s_2 \mid z_2)$. Together it follows that:
\begin{align}
    I(s_1; s_2; z_1; z_2) &= I(s_1; s_2; z_1) - I(s_1; s_2; z_1 \mid z_2) \\
     &= I(s_1; s_2) - I(s_1; s_2 \mid z_2) \\
     &= I(s_1; s_2; z_2) \\
     &= I(s_1; s_2) > 0
\end{align}
On the other hand, we know that $0 = H(s_1 \mid z_1) = H(s_1 \mid z_1; z_2) + I(s_1, z_2 \mid z_1)$ and therefore $I(s_1, z_2 \mid z_1) = 0$. Therefore, the interaction information $I(s_1; z_2; z_1) = I(s_1; z_2) - I(s_1; z_2 \mid z_1) = I(s_1; z_2) \geq 0$. At the same time, we assumed that $I(z_1; z_2) = 0$ and hence $I(z_1; z_2; s_1) + I(z_1; z_2 \mid s_1) = 0$, which shows that $I(z_1; z_2; s_1) \leq 0$. Together, we see that $I(z_1; z_2; s_1) = I(s_1; z_2) = 0$.

Now we can decompose $I(s_1; s_2; z_1; z_2)$ in a different way: $I(s_1; s_2; z_1; z_2) = I(s_1; z_1; z_2) - I(s_1; z_1; z_2 \mid s_2)$. We know that $I(s_1; z_1; z_2) = I(s_1; z_2)$ and therefore $I(s_1; z_1; z_2 \mid s_2) = I(s_1; z_2 \mid s_2) > 0$ and that $I(s_1; z_1; z_2) = 0$. Therefore, it follows that:
\begin{align}
    I(s_1; s_2; z_1; z_2)
    &=
    I(s_1; z_1; z_2) - I(s_1; z_1; z_2 \mid s_2) \\
    &=
    0 - I(s_1; z_2 \mid s_2) \\
    &\leq 0
\end{align}
which is a contradiction with $I(s_1; s_2; z_1; z_2) = I(s_1; s_2) > 0$. Therefore, if $I(s_1; s_2) > 0$ and $I(z_1; z_2) = 0$, it must hold that $I(z_k; s_k) < H(s_k)$ for at least one $k$, which we wanted to show.
\end{proof}

\end{document}